\documentclass[10pt,journal,compsoc]{IEEEtran}

\ifCLASSOPTIONcompsoc
  \usepackage[nocompress]{cite}
\else
  \usepackage{cite}
\fi
\usepackage{mathtools}
\usepackage{dsfont}
\usepackage{amssymb}
\usepackage{amsmath}
\usepackage{amstext}
\usepackage{makecell}
\usepackage{multirow}
\usepackage{textcomp}
\usepackage{array}
\usepackage{subcaption}
\newcolumntype{?}{!{\vrule width 1pt}}

\begin{document}

\title{Structured Label Inference for \\ Visual Understanding}

\author{
Nelson Nauata, 
Hexiang Hu,
Guang-Tong Zhou,
Zhiwei Deng,\\
Zicheng Liao and Greg Mori

}

\markboth{IEEE TRANSACTIONS ON PATTERN ANALYSIS AND MACHINE INTELLIGENCE,~Vol.~X, No.~Y, MONTH Z}%
{Shell \MakeLowercase{\textit{et al.}}: Bare Advanced Demo of IEEEtran.cls for IEEE Computer Society Journals}

\IEEEtitleabstractindextext{%
\begin{abstract}
Visual data such as images and videos contain a rich source of structured semantic labels as well as a wide range of interacting components. Visual content could be assigned with fine-grained labels describing major components, coarse-grained labels depicting high level abstractions, or a set of labels revealing attributes. Such categorization over different, interacting layers of labels evinces the potential for a graph-based encoding of label information. In this paper, we exploit this rich structure for performing graph-based inference in label space for a number of tasks: multi-label image and video classification and action detection in untrimmed videos. We consider the use of the Bidirectional Inference Neural Network (BINN) and Structured Inference Neural Network (SINN) for performing graph-based inference in label space and propose a Long Short-Term Memory (LSTM) based extension for exploiting activity progression on untrimmed videos. The methods were evaluated on (i) the Animal with Attributes (AwA), Scene Understanding (SUN) and NUS-WIDE datasets for multi-label image classification, (ii) the first two releases of the YouTube-8M large scale dataset for multi-label video classification, and (iii) the THUMOS'14 and MultiTHUMOS video datasets for action detection. Our results demonstrate the effectiveness of structured label inference in these challenging tasks, achieving significant improvements against baselines.

\end{abstract}

}

\maketitle

\IEEEdisplaynontitleabstractindextext
\IEEEpeerreviewmaketitle

\ifCLASSOPTIONcompsoc
\IEEEraisesectionheading{\section{Introduction}\label{sec:introduction}}
\else
\section{Introduction}
\label{sec:introduction}
\fi



\IEEEPARstart{V}{isual} content is a rich source of high-dimensional structured data, with a wide range of interacting components at varying levels of abstractions. With the proliferation of large-scale image \cite{imagenet, Chua:2009:NRW, 6571196, 5539970} and video \cite{youtube8m, THUMOS14, yeung2015every} datasets, advances in visual understanding were facilitated for the exploration and enhancement of intelligent reasoning techniques for modelling structured concepts. In this paper, we exploit these rich structures for modelling concept interactions in a number of different tasks and levels of complexity: multi-label image classification, multi-label video classification and action detection in untrimmed videos.

\begin{figure}
\centering
\includegraphics[width=\columnwidth]{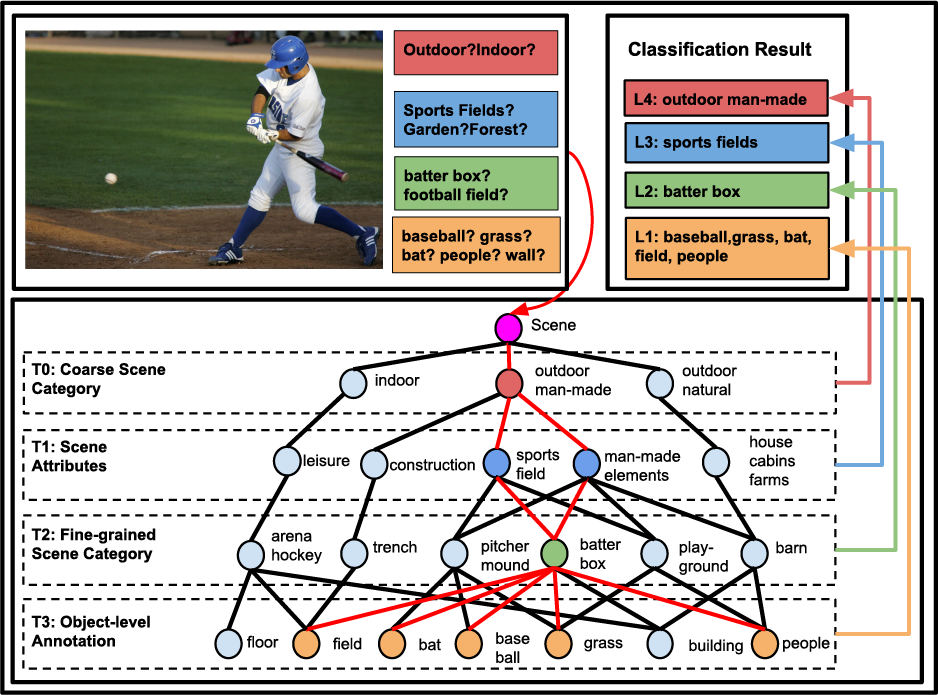}
\caption{This image example has visual concepts at various levels,
from sports field at a high level to baseball and person at lower
levels. Our model leverages label relations and jointly predicts layered
visual labels from an image using a structured inference neural
network. In the graph, colored nodes correspond to the labels
associated with the image, and red edges encode label relations.}
\label{fig:binn}
\end{figure}

Standard image classification is a fundamental problem in computer vision -- assigning category labels to images. It can serve as a building block for many different computer vision tasks including object detection, visual segmentation, scene parsing and concept localization. Successful deep learning approaches \cite{Krizhevsky:2017:ICD:3098997.3065386, Sermanet_overfeat:integrated, Simonyan14c, 43022} typically assume labels to be semantically independent and adapt either a multi-class or binary classifier to target labels. In recent work \cite{labelgraph, 7410495}, deep learning methods that take advantage of label relations have been proposed to improve classification performance. However, in realistic settings, these label relationships could form a complicated graph structure. Take Figure \ref{fig:binn} as an example. Various levels of interpretation could be formed to represent such visual content. This image of a \textit{baseball} scene could be described as an \textit{outdoor} image at coarse level, or with a more concrete concept such as \textit{sports field}, or with even more fine-grained labels such as \textit{batter's box} and objects such as \textit{grass}, \textit{bat}, \textit{person}.

Models that incorporate semantic label relationships could be utilized to generate better classification results. The desiderata for these models include the ability to model label-label relations such as positive or negative correlation, respect multiple concept layers obtainable from sources such as WordNet, and to handle partially observed label data – given a subset of accurate labels for this image, infer the remaining missing labels. 

In our previous work \cite{inn}, we developed a structured inference neural network that permits modeling complex relations between labels, ranging from hierarchical to within-layer dependencies. We achieve this by defining a network in which a node is activated if its corresponding label is present in an image. We introduce stacked layers among these label nodes. These encode layer-wise connectivity among label classification scores, representing dependencies from top-level coarse labels to bottom-level fine-grained labels. Activations are propagated bidirectionally and asynchronously on the label relation graph, passing information about the labels within or across concept layers to refine the labeling for the entire image.

The similarity between multi-label classification on images and videos suggests exploitation of structured data to be beneficial for both tasks. As demonstrated in \cite{DBLP:journals/corr/NauataSM17}, our method extends beyond its applicability on images and is robust to higher dimensional structured data such as videos. A more challenging problem than multi-label video classification consists of handling a sequential input of frames and inferring the corresponding sequence of dense annotations. Our exploration of this setting for the task of dense action detection \cite{yeung2015every} is two-fold: performing static frame structured inference and spatio-temporal structured inference.

The static frame inference can be reduced to a standard image classification problem.  However, spatio-temporal structured inference requires modelling cross-temporal relationships between labels. A natural way of extending \cite{DBLP:journals/corr/NauataSM17} to support this feature is allowing communication between concept layers in the hierarchical structure for forward propagating learned label correlations and exploring labels' progression on untrimmed videos. We achieve this by enriching the hidden state of Long Short-Term Memory (LSTM) \cite{lstm} units with extracted information from our structured inference method along the frame sequences.

The contribution of this paper is binding together the proposed model presented in \cite{inn} for performing hierarchical inference on image datasets (AwA \cite{6571196} and SUN397 \cite{5539970}) and its video extension previously implemented in \cite{DBLP:journals/corr/NauataSM17}. In addition, we include novel results for the two recent releases of Youtube-8M \cite{youtube8m}, including partially observed labels and propose a temporal extension for the bidirectional and structured inference models, demonstrating that adding cross-temporal information in label space (i.e. propagation across concept layers) provides superior performance against a traditional technique for incorporating temporal dependencies (i.e. LSTM). The validation of the proposed models was carried out on THUMOS \cite{THUMOS14} and MultiTHUMOS \cite{yeung2015every} for the task of action detection.

This paper is organized as follows. Section \ref{background} presents prior knowledge, covering previous related work. In Section \ref{methods}, we bind together the bidirectional and structured inference models formulated in \cite{inn} with its video extension presented in \cite{DBLP:journals/corr/NauataSM17} and derive the proposed formulation for temporal extension of our bidirectional and structured inference methods. Sections \ref{img_task} and \ref{vid_task} present the work done on multi-label image and video classification respectively. Section \ref{det_task} describes the work done on the action detection task. We conclude in Section \ref{conclusion} with a brief discussion.

\section{Related Work}\label{background}


Structured labeling information has been incorporated in numerous frameworks throughout previous literature. In this section, we review some prominent related works and the algorithms implemented for the various tasks studied in this paper.

\subsection{Label Relations and External Structured Concepts}
The incorporation of label relations is often explored by modelling graphical structures on the training data (e.g. \cite{NIPS2011_4212}) or constraining a loss function for jointly predicting structured labels \cite{Taskar:2003:MMN:2981345.2981349, Tsochantaridis:2005:LMM:1046920.1088722}. This work investigates the ability to model label relations such as positive and negative correlations, building on top of the bidirectional and structured inference neural network previously proposed in \cite{inn}. 

When external structured concepts are available it is advantageous to incorporate them to conduct traditional supervised approaches. For instance, Grauman \emph{et al.} \cite{NIPS2011_4250} and Hwang \emph{et al.} \cite{hwang2012semantic} exploit the WordNet taxonomy for learning a discriminative tree of metrics of visual representations hierarchically structured. Johnson \emph{et al.} \cite{Johnson2015ICCV} and McAuley and Leskovec \cite{conf/eccv/McAuleyL12} exploited social-network metadata, taking into account its interdependencies for applying structured learning and enhancing image classification against methods relying solely on image content.

\subsection{Multi-task Joint Learning}
The intuition behind structured label prediction is closely related to multi-task learning, with the distinction that multiple correlated tasks are jointly estimated. Common jointly modeled tasks include segmentation and detection \cite{1467244, 4270067}, segmentation and pose estimation \cite{Kohli2008}, or segmentation and object classification \cite{Leibe04combinedobject}. An emerging topic of joint learning lies in image understanding and text generation by leveraging intra-modal correspondences between visual data and human language \cite{KongCVPR14, 7534740}. 

The method presented in \cite{inn} might also be extended to multi-task learning. A natural way of achieving this is considering each concept layer to be a different task, whose labels do not necessarily contain a layered structure.  Thus, the existing multi-task learning methods might also benefit from  prior knowledge of intra-task relations.

\subsection{Deep Visual Recognition}
The use of deep learning for applications in videos has advanced in lockstep with the field's success in images. A traditional deep learning framework for extracting spatial descriptions comprises using convolutional neural networks (CNNs) such as AlexNet~\cite{Krizhevsky:2017:ICD:3098997.3065386}, VGG-16 \cite{Simonyan14c} or InceptionV3~\cite{inceptionv3}, pre-trained on a diverse and large image collection (e.g. ImageNet \cite{imagenet}). Recently, deep residual networks~\cite{he2016deep} built on these successes, and pushed the image classification performance of deep CNNs to impressive performance levels on benchmark datasets. In this paper, we leverage those recent progresses for structured label inference.

\subsection{Video Classification}

Traditional approaches for video recognition \cite{HengWang:2011:ARD:2191740.2192078, 6751553} build classifiers on top of hand-crafted features such as Histograms of Oriented Gradients (HOG), Histograms of Optical Flow (HOF) and Motion Boundary Histograms (MBH). On the contrary, common deep learning strategies for improving recognition consist of automatically learning descriptors that capture discriminative appearance and motion features via implementation of spatio-temporal networks \cite{6909619, 6165309, 10.1007/978-3-642-25446-8_4} or exploitation of temporal dependencies using recurrent neural networks \cite{beyond_horts_nippets, lrcn}. Instead our approach intends to leverage structured label information for performing stronger video-level label predictions. For this reason, we adopt the simple frame-level feature aggregation from \cite{youtube8m}, which consists of average pooling spatial features across untrimmed YouTube videos \cite{youtube8m} for performing video classification. 



\subsection{Action Detection}

More challenging than performing video-level recognition is localizing concepts at a frame-level. In this paper, we address the problem of dense multi-label action detection.  We utilize a recent dataset~\cite{yeung2015every}, which requires inferring multiple frame-level labels across videos. A classic approach before deep learning \cite{4409011} tackles the problem of detection by matching volumetric representations of events against oversegmented videos. Aligned more closely to our approach are \cite{7780585, Montes_2016_NIPSWS, 7780583}, which analyze label progression, operating on short clips or single frames for inferring actions for each time step using recurrent neural networks. In contrast, we consider label progression over a structured collection of labels. 




\subsection{Recurrent Temporal Modeling}\label{lstm}
Early attempts of modelling time-sequential images in computer vision rely on hidden Markov models for obtaining representations in order to recognize action classes considering evolution of actions or even multiple view-points \cite{223161, 4270156, Shi2011}. A stronger presence of deep learning in recent years along with its eminent success in the field, shifted attention to recurrent neural network (RNN) approaches for modelling time dependent correlations in sequential data. In particular for frame sequences, LSTM based models have been extensively explored recently \cite{beyond_horts_nippets, lrcn, yeung2015every} due to their capacity of efficiently transmitting information across time. In this paper, we leverage such successes of temporal modelling techniques for extending our structured label inference framework across time. 





\begin{figure*}[htbp]
  \includegraphics[width=\textwidth]{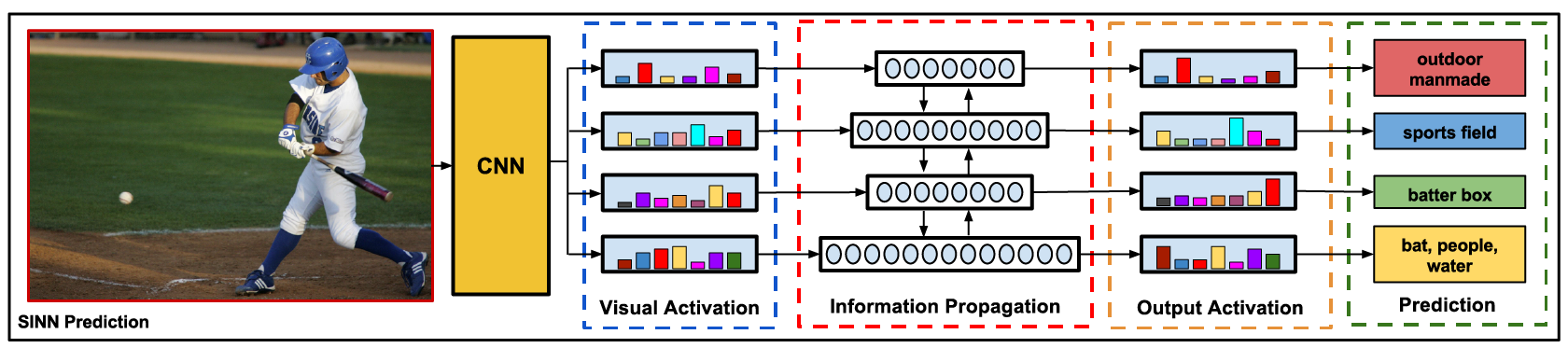}
  \caption{ The label prediction pipeline. Given an input image, we extract CNN features at the last fully connected layer as activation (in blue box) at different visual concept layers. We then propagate the activation information in a label (concept) relation graph through our structured inference neural network (in red box). The final label prediction (in green box) is made from the output activations (in yellow box) obtained after the inference process.}
  \label{sinn_framework}
\end{figure*}

\section{Structured Inference Framework}\label{methods}

Our model jointly classifies images or videos in a layered label space with external label relations. The goal is to leverage the label relations to improve inference over the layered visual concepts and extend our approach for modelling temporal dependencies for action detection.

We build our model on top of a state-of-the-art deep learning platform: given a collection of training images $\mathcal{I}$ (i.e. $\{I^i\}^{|\mathcal{I}|}_{i=1}$) or a collection of videos $\mathcal{V}$ (i.e. $\{V^j\}^{|\mathcal{V}|}_{j=1}$), where the j-th video is composed by frames $\{F^{j}_{k}\}_{k=1}^{|V^j|}$, we first reduce spatial dimensions by extracting CNN features as visual activations. For example, a feature vector $\mathbf{x}^i$ for the i-th image in the collection is obtained by $CNN(I^i)$, or a feature vector $\mathbf{x}^{j}_{k}$ for the j-th video and k-th frame is obtained by applying $CNN(F^{j}_{k})$.

In the particular case of inputting single images or frames (i.e. $I^{i}$ or $F^{j}_{k}$), spatial representations $\mathbf{x}^{i}$ or $\mathbf{x}^{j}_{k}$, can be fed directly to our model. For an entire video $\{F^{j}_{k}\}_{k=1}^{|V^j|}$, we deal with inputs in either of two ways: providing a summarized feature vector $\mathbf{\bar{x}}^j$ for the entire video or a sequence of feature vectors $\{\mathbf{x}_{k}^{j}\}_{k=1}^{|V_j|}$ extracted from the corresponding input frames. For video classification, we apply the former case by including an intermediate averaging pooling across the entire video \cite{youtube8m} and the latter case serves as input to our temporal extension model.

In Section \ref{binn_sec} and Section \ref{sinn_sec}, we formulate our models for tagging with multiple labels $\mathbf{y}^{i}$ a single image $I^i$, or with multiple video-level labels $\mathbf{y}^{j}$ an entire video $V^j$. In Section \ref{temp_ext_sec}, we describe our proposed temporal extension method for outputting multiple per-frame actions $\{\mathbf{y}_{k}^{j}\}_{k=1}^{|V_j|}$ in a video $V_j$, using a temporal sequence of spatial feature vectors $\{\mathbf{x}_{k}^{j}\}_{k=1}^{|V_j|}$ as input. In Section \ref{pred_sec}, we present our approach for performing structured predictions and including partial observations in our models.


\subsection{Bidirectional Inference Neural Network (BINN)}\label{binn_sec}

Our model is inspired by the recent success of RNNs \cite{NIPS2008_3449, DBLP:conf/interspeech/2014}, which make use of dynamic sequential information in learning. RNNs are called recurrent models because they perform the same computation for every time step, with the input dependent on the current inputs and previous outputs. Applying a similar idea to our layered label prediction problem: we consider each concept layer as an individual time step, and model the label relations within and across concept layers in the recurrent learning framework.

The learning of structured label relations is seen as a hierarchical distribution of labels in our architecture, where each level is defined as a concept layer and represents the degree of granularity for the label space. In the implementation for the BINN, the labels are separated into a set $\mathcal{M}$ of concept layers, with varying granularity, totalling $m$ layers. For example, a coarse-grained label for a scene could be \textit{outdoors}, whereas a fine-grained label would be \textit{tree}.


From now on, we denote a single input feature vector $\mathbf{x} \in \mathbb{R}^D$ with ground-truth labels as a binary vector $\mathbf{t}^{\ell}$ at concept layer $\ell \in \mathcal{M}$, where $t^{\ell}_ k = 1$, if the k-th concept is present and $t^{\ell}_ k = 0$, otherwise. This feature vector will be treated as input separately for each concept layer.
The concept layers are represented as activations obtained by performing inference in the graph -- for each concept layer $\ell$, we have an activation vector $\mathbf{a}^{\ell} \in \mathbb{R}^{n_{\ell}}$ associated with the labels at concept layer $\ell$, where $n_{\ell}$ is the number of labels at concept layer $\ell$. In order to perform inference, the dimension of the input $\mathbf{x}$ should be regressed to the label space. Thus, the input $\mathbf{x}^{\ell}$ for each concept layer $\ell$ and an input feature vector $\mathbf{x}$ is given by:
\begin{align}\label{eqn:binn_in}
	\mathbf{x}^{\ell} = W^{\ell} \cdot \mathbf{x} + \mathbf{b}^{\ell},
\end{align}
where $W^{\ell} \in \mathbb{R}^{n_{\ell} \times D}$ and $\mathbf{b}^{\ell} \in \mathbb{R}^{n_{\ell} \times 1}$ are learnable parameters.


The bidirectional message passing consists of two steps: a top-down inference and a bottom-up inference. The former captures inter-layer and intra-layer label relations in the top-down direction computing intermediate activations represented by $\overrightarrow{\mathbf{a}}^{\ell}$. The latter performs the same computation in the bottom-up direction and are represented by $\overleftarrow{\mathbf{a}}^{\ell}$. The aggregation parameters $\overrightarrow{U^{\ell}}$ and $\overleftarrow{U^{\ell}}$ are defined for combining both directions and obtaining final activations $\mathbf{a}^{\ell}$ for concept layer $\ell$.

The entire formulation for a feature vector $\mathbf{x}$, after obtaining $\mathbf{x}^{\ell}$ for all concept layers using Eq. \eqref{eqn:binn_in}, is the following:
\begin{align}
		\overrightarrow{\mathbf{a}}^{\ell} &= \overrightarrow{V}^{\ell-1,\ell} \cdot \overrightarrow{\mathbf{a}}^{\ell-1} + \overrightarrow{H}^{\ell} \cdot \mathbf{x}^{\ell} + \overrightarrow{\mathbf{b}}^{\ell}, \label{eqn:binn_eqn_1} \\
		\overleftarrow{\mathbf{a}}^{\ell} &= \overleftarrow{V}^{\ell+1,\ell} \cdot \overleftarrow{\mathbf{a}}^{\ell+1} + \overleftarrow{H}^{\ell} \cdot \mathbf{x}^{\ell} + \overleftarrow{\mathbf{b}}^{\ell}, \label{eqn:binn_eqn_2}\\ 
		\mathbf{a}^{\ell} &= \overrightarrow{U}^{\ell} \cdot \overrightarrow{\mathbf{a}}^{\ell} + \overleftarrow{U}^{\ell} \cdot \overleftarrow{\mathbf{a}}^{\ell} + \mathbf{b}^{a,\ell},  \label{eqn:binn_eqn_3}
\end{align}
where $V^{i, j} \in \mathbb{R}^{n_j \times n_i}$, $H^i \in \mathbb{R}^{n_i \times n_i}$, $U^i \in \mathbb{R}^{n_i}$, and $\mathbf{b}^{a, i}, \mathbf{b}^{i} \in \mathbb{R}^{n_i}$ are all learnable parameters. The V's and H's weights capture the inter-layer and intra-layer dependencies, respectively. Since these parameters exhaust all pairwise relationships between labels, this step can be thought of as propagating activations across a fully-connected directed label graph.

In order to obtain concept-specific probabilities for making predictions, the activations $a^{\ell}_k$ are passed through a sigmoid function (i.e. $\sigma(z)=\frac{1}{1+e^{-z}}$), for concepts $k \in \ell$ and concept layers $\ell \in \mathcal{M}$, yielding normalized activations $y^{\ell}_k=\sigma(a^{\ell}_k)$, used as probabilities. To learn the layer parameters, the model is trained end-to-end with backpropagation, minimizing logistic cross-entropy loss for a given batch of activations $\{a^{\ell}_{k}\}$, as follows:
\begin{align}
		E(\{a^{\ell}_{k}\}) &= -\sum_{i=1}^{N}\sum\limits_{\ell=1}^{m} \sum\limits_{k=1}^{n_{\ell}} \bigg( t^{\ell}_{k, i} \cdot \log \big( \sigma (a^{\ell}_{k,i}) \big) \label{eqn:loss_func} \\
			 &+ (1-t^{\ell}_{k, i} ) \cdot \log \big( 1 - \sigma (a^{\ell}_{k, i}) \big) \bigg)\notag,
\end{align}
where $N$ denotes the batch size selected as a hyperparameter, $t^{\ell}_{k, i}$ and $\sigma (a^{\ell}_{k, i})$ correspond to the ground-truth label and output score, respectively, for the k-th concept in layer $\ell$ for the i-th sample in the batch.



\subsection{Structured Inference Neural Network (SINN)} \label{sinn_sec}

The fully connected bidirectional model is capable of representing all types of label relations. In practice, however, it is hard to train a model on limited data due to the large number of free parameters. To avoid this problem, we use a structured label relation graph to restrict information propagation.

\begin{figure}[htbp]
  \includegraphics[width=0.5\textwidth]{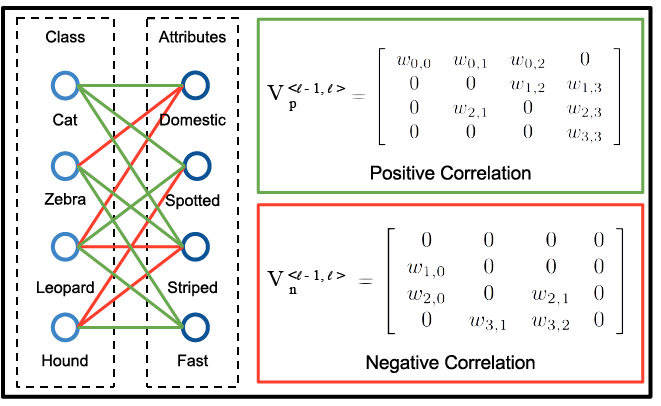}
  \caption{ An example showing the model parameters V\textsubscript{p} and V\textsubscript{n} between the animal layer and the attribute layer. Green edges in the graph represent positive correlation, and red edges represent negative correlation.}
  \label{sinn_structure}
\end{figure}

We use structured label relations of positive correlation and negative correlation as prior knowledge to refine the model. The intuition is as follows: since we know that \textit{office} is an \textit{indoor} scene, \textit{beach} is an \textit{outdoor} scene, and \textit{indoor} and \textit{outdoor} are mutually exclusive, a high score on \textit{indoor} should increase the probability of label \textit{office} and decrease the probability of label \textit{beach}. Labels that are not semantically related, e.g. \textit{motorcycle} and \textit{shoebox}, should not affect each other. The structured label relations can be obtained from sources such as semantic taxonomies, or by parsing WordNet relations \cite{Miller:1995:WLD:219717.219748}. We introduce the notation $V_p$, $V_n$, $H_p$ and $H_n$ to explicitly capture structured label relations in between and within concept layers, where the subscripts \textit{p} and \textit{n} indicate positive and negative correlation, respectively. These model parameters are masked matrices capturing the label relations. Instead of learning full parametrized matrices $V_p$, $V_n$, $H_p$ and $H_n$, we freeze some elements to be zero if there is no semantic relation between the corresponding labels. For example, $V_p$ models the positive correlation in between two concept layers: only the label pairs that have positive correlation have learnable model parameters, while the rest are zeroed out to remove potential noise. A similar setting goes to $V_n$, $H_p$ and $H_n$. Figure \ref{sinn_structure} shows an example positive correlation graph and a negative graph between two layers.

To implement the positive and negative label correlation, we implement the following structured message passing process:

\begin{align}
		\overrightarrow{\mathbf{a}}^\ell &=  \gamma(\overrightarrow{V}^{\ell-1,\ell}_p \cdot \overrightarrow{\mathbf{a}}^{\ell-1}) + \gamma(\overrightarrow{H}^{\ell}_p \cdot \mathbf{x}^\ell) \label{eqn:sinn_eqn_1} \\
		&\quad-\gamma(\overrightarrow{V}^{\ell-1,\ell}_n \cdot \overrightarrow{\mathbf{a}}^{\ell-1}) - \gamma(\overrightarrow{H}^{\ell}_n \cdot \mathbf{x}^{\ell}) + \overrightarrow{\mathbf{b}}^{\ell},\notag  \\
		\overleftarrow{\mathbf{a}}^{\ell} &=  \gamma(\overleftarrow{V}^{\ell+1,\ell}_p \cdot \overleftarrow{\mathbf{a}}^{\ell+1}) + \gamma(\overleftarrow{H}^{\ell}_p \cdot \mathbf{x}^{\ell})\label{eqn:sinn_eqn_2}\\
		&\quad-\gamma(\overleftarrow{V}^{\ell+1,\ell}_n \cdot \overleftarrow{\mathbf{a}}^{\ell+1}) - \gamma(\overleftarrow{H}^{\ell}_n \cdot \mathbf{x}^{\ell}) + \overleftarrow{\mathbf{b}}^{\ell},\notag\\
		\mathbf{a}^{\ell} &= \overrightarrow{U}^{\ell} \cdot \overrightarrow{\mathbf{a}}^{\ell} + \overleftarrow{U}^{\ell} \cdot \overleftarrow{\mathbf{a}}^{\ell} + \mathbf{b}^{a,\ell}.  \label{eqn:sinn_eqn_3}
\end{align}

\begin{figure*}[htbp]
  \includegraphics[width=\textwidth]{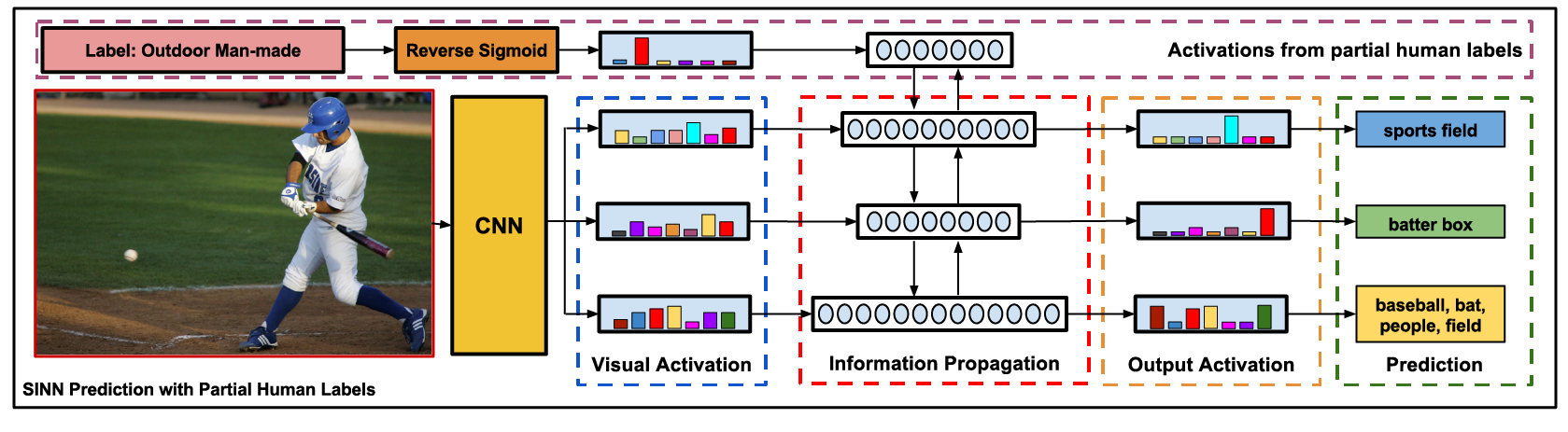}
  \caption{ The label prediction pipeline with partial observation. The pipeline is similar to Figure \ref{sinn_framework} except that we now have a partial observation that this image is outdoor man-made. The SINN is able to take the observed label into consideration and improve the label predictions in the other concept layers.}
  \label{label_prediction}
\end{figure*}

Here $\gamma(.)$ stands for a ReLU activation function. It is essential for SINN as it enforces that activations from positive correlation always make positive contribution to output activation and keeps activations from negative correlation as negative contribution (notice the minus signs in Eqs. \eqref{eqn:sinn_eqn_1} and \eqref{eqn:sinn_eqn_2}). To learn the model parameters V's, H's, and U's, we optimize the cross-entropy loss in Eq. \eqref{eqn:loss_func}.

\subsection{Modeling Cross-temporal Relations}\label{temp_ext_sec}

In order to model temporal dependencies between frames benefiting from the inclusion of structured label relationships, we extend BINN to allow message passing across time. Since this temporal extension is built on top of a LSTM model (Section \ref{lstm}) and benefits from bidirectional inference in label space, we refer to it as bidirectional inference LSTM (biLSTM) from now on. Thus, considering an input sequence $\{\mathbf{x}_{t}\}_{t=1}^{T}$, we can rewrite Eq. \eqref{eqn:binn_in} as follows:

\begin{align}\label{eqn:ilstm_inn}
	\mathbf{x}^{\ell}_t = W^{\ell} \cdot \mathbf{x}_t + \mathbf{b}^{\ell},
\end{align}
where $W^{\ell}$ and $\mathbf{b}^{\ell}$ are parameters shared across time.

Using the same notation for the bidirectional message passing presented in Section \ref{binn_sec}, Eq. \eqref{eqn:binn_eqn_1}, \eqref{eqn:binn_eqn_2} and \eqref{eqn:binn_eqn_3} can be extended as follows:

\begin{align}
		\overrightarrow{\mathbf{a}}^{\ell}_t &= \overrightarrow{V}^{\ell-1,\ell} \cdot \overrightarrow{\mathbf{a}}^{\ell-1}_t + \overrightarrow{H}^{\ell} \cdot \mathbf{x}^{\ell}_t + \overrightarrow{\mathbf{b}}^{\ell}, \\ 
		\overleftarrow{\mathbf{a}}^{\ell}_t &= \overleftarrow{V}^{\ell+1,\ell} \cdot \overleftarrow{\mathbf{a}}^{\ell+1}_t + \overleftarrow{H}^{\ell} \cdot \mathbf{x}^{\ell}_t + \overleftarrow{\mathbf{b}}^{\ell}, \\
		\mathbf{a}^{\ell}_t &= \overrightarrow{U}^{\ell} \cdot \overrightarrow{\mathbf{a}}^{\ell}_t + \overleftarrow{U}^{\ell} \cdot \overleftarrow{\mathbf{a}}^{\ell}_t + \mathbf{b}^{a,\ell} \label{eq:biLstm_agg},
\end{align}
where $V^{i, j} \in \mathbb{R}^{n_j \times n_i}$, $H^i \in \mathbb{R}^{n_i \times n_i}$, $U^i \in \mathbb{R}^{n_i}$, and $\mathbf{b}^{a, i}, \mathbf{b}^{i} \in \mathbb{R}^{n_i}$ are shared across time.

Next, we combine the output $\mathbf{a}^{\ell}_t$ with the hidden state of a LSTM after each time step. Assuming $\sigma(\cdot)$ and $tanh(\cdot)$ are the sigmoid and hyperbolic tangent functions, we rewrite the LSTM unit definition from \cite{yeung2015every}, as follows:

\begin{align}
		\mathbf{i}^{\ell}_t &= \sigma(W^{\ell}_{xi} \cdot \mathbf{x}_t + W^{\ell}_{hi} \cdot \mathbf{h}^{\ell}_{t-1} + \mathbf{b}^{\ell}_i), \\
		\mathbf{f}^{\ell}_t &= \sigma(W^{\ell}_{xf} \cdot \mathbf{x}_t + W^{\ell}_{hf} \cdot \mathbf{h}^{\ell}_{t-1} + \mathbf{b}^{\ell}_f), \\
		\mathbf{o}^{\ell}_t &= \sigma(W^{\ell}_{xo} \cdot \mathbf{x}_t + W^{\ell}_{ho} \cdot \mathbf{h}^{\ell}_{t-1} + \mathbf{b}^{\ell}_o), \\
        \mathbf{g}^{\ell}_t &= \tanh(W^{\ell}_{xc} \cdot \mathbf{x}_t + W^{\ell}_{hc} \cdot \mathbf{h}^{\ell}_{t-1} + \mathbf{b}^{\ell}_c), \\
		\mathbf{c}^{\ell}_t &= \mathbf{f}_t * \mathbf{c}^{\ell}_{t-1} + \mathbf{i}^{\ell}_t * \mathbf{g}^{\ell}_t, \\
		\mathbf{h}^{\ell}_t &= \mathbf{o}^{\ell}_t * \tanh(\mathbf{c}^{\ell}_t),
\end{align}
where $*$ denotes point-wise multiplication. This formulation is equivalent to appending one LSTM unit for each concept layer $\ell$. Intuitively, we allow the model to accumulate information about each concept layer separately, which empirically provided better results than assigning one memory unit for all concept layers.

The concept layer's activations $\mathbf{a}^{\ell}_t$ are then combined with its memory unit's hidden state $\mathbf{h}^{\ell}_t$ at time step $t$, as follows:
\begin{align}\label{agg_step}
		\mathbf{y}^{\ell}_t = \sigma(M^a \cdot \mathbf{a}^{\ell}_t + M^h \cdot \mathbf{h}^{\ell}_t + \mathbf{b}^{a,h}),
\end{align}
where $M^a$, $M^h$ and $\mathbf{b}^{a,h}$ are parameters shared across time, $\mathbf{y}^{\ell}_t$ corresponds to normalized activations for concept layer $\ell$ at time step $t$ used as confidence scores and $\sigma(\cdot)$ is the sigmoid function.

Similarly, we extend the SINN model to allow the flow of information across time as well, by rewriting Eqs. \eqref{eqn:sinn_eqn_1} and \eqref{eqn:sinn_eqn_2}, as follows:
\begin{align}
	\overrightarrow{\mathbf{a}}^{\ell}_t &= \gamma(\overrightarrow{V}^{\ell-1,\ell}_p \cdot \overrightarrow{\mathbf{a}}^{\ell-1}_t) + \gamma(\overrightarrow{H}^{\ell}_p \cdot \mathbf{x}^{\ell}_t)\label{eqn:siLstm_eqn_1} \\
	&\quad-\gamma(\overrightarrow{V}^{\ell-1,\ell}_n \cdot \overrightarrow{\mathbf{a}}^{\ell-1}_t) - \gamma(\overrightarrow{H}^{\ell}_n \cdot \mathbf{x}^{\ell}_t) + \overrightarrow{\mathbf{b}}^{\ell},\notag  \\
	\overleftarrow{\mathbf{a}}^{\ell}_t &= \gamma(\overleftarrow{V}^{\ell+1,\ell}_p \cdot \overleftarrow{\mathbf{a}}^{\ell+1}_t) + \gamma(\overleftarrow{H}^{\ell}_p \cdot \mathbf{x}^{\ell}_t)\label{eqn:siLstm_eqn_2}\\
	&\quad-\gamma(\overleftarrow{V}^{\ell+1,\ell}_n \cdot \overleftarrow{\mathbf{a}}^{\ell+1}_t) - \gamma(\overleftarrow{H}^{\ell}_n \cdot \mathbf{x}^{\ell}_t) + \overleftarrow{\mathbf{b}}^{\ell},\notag
\end{align}
using the same notation as for the previously formulated models. Note that the aggregation equation stays the same as in Eq. \eqref{eq:biLstm_agg}. In order to distinguish this model from the biLSTM, we refer to it as structured inference LSTM (siLSTM). The output activations $\mathbf{y}^{\ell}_t$ serve as confidence scores for each concept at layer $\ell$ being assigned to the frame at time step $t$ and is used for outputting the detections. 

\subsection{Prediction Framework}\label{pred_sec}
Now we introduce the method of predicting labels during test time. As the model is trained with multiple concept layers, it is straightforward to recognize a label at each concept layer for the provided test sample. This mechanism is called label prediction \textit{without observation} (the default pipeline shown in Figure \ref{sinn_framework}).

Another interesting application is to make predictions with \textit{partial observations} -- we want to predict labels in one concept layer given labels in another concept layer. Figure \ref{label_prediction} illustrates the idea. Given an image shown in the left side of Figure \ref{label_prediction}, we have more confidence to predict it as \textit{batter's box} once we know it is an \textit{outdoor} image with attribute \textit{sports field}.

To make use of the partially observed labels in our SINN framework, we need to transform the observed binary labels into soft activation scores for SINN to improve the label prediction on the target concept layers. Recall that SINN minimizes cross-entropy loss which applies sigmoid functions on activations to generate label confidences. Thus, we reverse this process by applying the inverse sigmoid function (logit) on the binary ground-truth labels to obtain activations. Formally, we define the k-th observed concept activation $a^{\ell}_k$ in concept layer $\ell$ obtained from a ground-truth label $t^{\ell}_k$ as:
\begin{equation}
a^{\ell}_k=
\begin{cases}
  log\frac{t^{\ell}_k+\epsilon}{1-(t^{\ell}_k+\epsilon)}, & \text{if}\ t^{\ell}_k = 0, \\
  log\frac{t^{\ell}_k-\epsilon}{1-(t^{\ell}_k-\epsilon)}, & \text{if}\ t^{\ell}_k = 1.
\end{cases}
\end{equation}
Note that we put a small perturbation $\epsilon$ on the ground-truth label $t^{\ell}_k$ for numerical stability. In our experiments, we set $\epsilon = 0.001$.





\begin{figure*}
\centering
\includegraphics[width=1.0\textwidth]{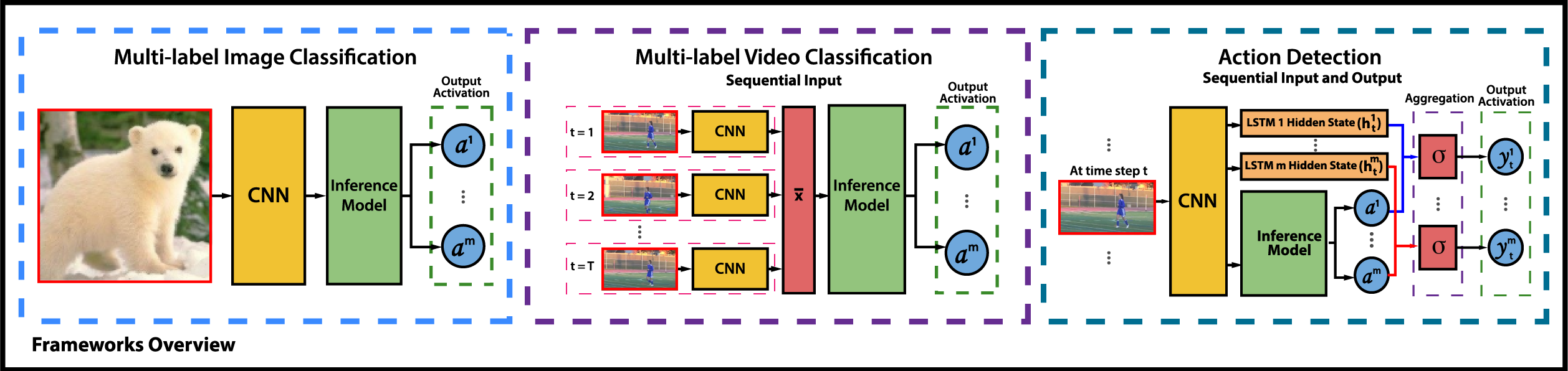}
\caption{Overview of the frameworks presented using the notation in Section \ref{methods}. (Left) The inference model is applied directly on a feature vector extracted by a CNN for performing multi-label image classification. (Center) The video-level representation $\bar{\mathbf{x}}$ is obtained by pooling per-frame feature vectors $\mathbf{x}_t$ (using a CNN) and fed to the inference model for performing multi-label video classification. (Right) Snapshot at time step $t$ for the action detection model, where per-frame representations $\mathbf{x}_t$ were extracted using a CNN, fed to the inference model and to $m$ concept-layer specific LSTM units at each time step $t$. The concept layer outputs $\{\mathbf{a}_{t}^{\ell}\}_{\ell=1}^{m}$ are combined (aggregation step denoted by $\sigma$ and formulated in Eq. \eqref{agg_step}) with the corresponding LSTM hidden states $\{\mathbf{h}_{t}^{\ell}\}_{\ell=1}^{m}$ for obtaining the final predictions $\{\mathbf{y}_{t}^{\ell}\}_{\ell=1}^{m}$. The output activations ($\mathbf{y}^{\ell}_{t}$) for the concept layer $\ell$ are interpreted as confidence scores for a given concept being assigned to the frame at time step $t$.}
\end{figure*}


\section{Multi-label Classification on Images} \label{img_task}
We start by describing our evaluation benchmarks for the task of multi-label classification on images (Section \ref{img_datasets}) as well as the corresponding metrics used for evaluation (Section \ref{img_metrics}). Next, we present our experimental setup and implementation details in Section \ref{img_exps} and \ref{img_details}, respectively. Finally, we report the results for this task in Section \ref{img_results}.

\subsection{Evaluation Benchmarks}\label{img_datasets}
\subsubsection{Animal with Attributes (AwA)}
The \textit{Animals with Attributes} (AwA) dataset was first introduced in \cite{6571196} for validating attribute-based classification performed in a zero-shot fashion. The AwA dataset consists of 30,475 images of animals queried from image search engines of \textit{Microsoft}, \textit{Google}, \textit{Yahoo} and \textit{Flickr}. The annotations comprises 50 animal classes and 85 and semantic attributes. In addition to the original annotations a 28-taxonomy-term layer was included \cite{inn} by parsing WordNet.

\subsubsection{Scene Understanding (SUN)}
The \textit{Scene Understanding} (SUN) database \cite{5539970} contains a total of 908 categories and 131,072 images. The final set of scene categories was obtained by parsing WordNet for collecting commonly found scenes. For the task of scene categorization only 397 categories (SUN397), for which there are at least 100 unique photographs were used for benchmarking. The SUN397 database contains an intrinsic hierarchical structure in label space defined by three levels of granularity annotated by Amazon's Mechanical Turk (AMT) workers. In increasing order of granularity SUN397 (from top to bottom) contains on the top layer 3 coarse categories, 16 general scene categories and 397 fine-grained scene categories on the bottom layer.

\subsubsection{Real-World Web Images (NUS-WIDE)}
The original NUS-WIDE \cite{Chua:2009:NRW} dataset consists of a large scale web image collection containing unique associated tags from Flickr checked against WordNet concepts. Over 300,000 images were randomly crawled together with their tags through the Flickr public API. After filtering duplicates, a total of 269,648 images remain, which include 5,018 associated unique tags and 81 ground-truth concepts.

\subsection{Evaluation Metrics}\label{img_metrics}
Here we define the metrics used for the multi-label image classification. All formulations are derived for a collection of images $\mathcal{I}$ with candidate classes $\mathcal{C}$. For each image sample $I \in \mathcal{I}$ with a ground-truth vector $\mathbf{t}_i$ are assigned multiple labels represented by a vector $\mathbf{y}_i$. 



\subsubsection{Multiclass Accuracy (MC Acc)}
The \textit{MC Acc} metric is a standard metric used for image classification and consists of averaging the per-class accuracies for all classes in the problem:

\begin{equation}
\it{MC \, Acc} =  \frac{1}{|\mathcal{C}|}\sum_{c \in \mathcal{C}}\frac{TP_{c}(\mathcal{I})}{|\mathcal{I}|},
\end{equation}
where $TP_{c}(\mathcal{I})$ denotes the true positives predictions for class $c$, considering all images in $\mathcal{I}$.


\subsubsection{Intersection over Union Accuracy (IoU Acc)}
Another widely used metric is the \textit{IoU Acc}, which is based on the \textit{Hamming distance} between predictions and the ground-truth labels:
\begin{equation}
\it{IoU \, Acc} =  \frac{1}{|\mathcal{I}|}\sum_{i = 1}^{|\mathcal{I}|} HammingDistance(\mathbf{y}_i, \mathbf{t}_i).
\end{equation}

\subsubsection{Average Precision (AP)}
In order to summarize the area under the precision-recall curve, the AP can be computed by averaging the precision weighed by increments in recall at each element in a list of annotation scores ranked by the model scores, as follows:
\begin{equation}\label{eqn:ap_equation}
\it{AP} =  \sum_{i = 1}^{N} P(i) \Delta R(i),
\end{equation}
where $N$ is the number of predictions, $P(i)$ and $\Delta R(i)$ are precision and increment recall at the i-th element in the ranked list.

\subsubsection{Per-label Mean Average Precision (mAP\textsubscript{L})}\label{lmap}
The mAP\textsubscript{L} is obtained by computing separately, the average precision (AP) for each label across all the samples in the target dataset. Given a list of annotation scores ranked by the model scores, we compute $AP_c$ for each class using Eq. \eqref{eqn:ap_equation} and calculate the mean over all classes:
\begin{equation}\label{eqn:ap_avgl}
\it{mAP_L} =  \frac{1}{|\mathcal{C}|}\sum_{c \in \mathcal{C}} AP_c.
\end{equation}


\subsubsection{Per Image Mean Average Precision (mAP\textsubscript{I})}
As opposed to mAP\textsubscript{L}, the mAP\textsubscript{I} considers the ranking of predictions for each image sample and compute the average across all samples in the dataset:
\begin{equation}\label{eqn:ap_avgi}
\it{mAP_I} =  \frac{1}{|\mathcal{I}|}\sum_{I \in \mathcal{I}} AP_I,
\end{equation}
where $AP_I$ is computed using Eq. \eqref{eqn:ap_equation} for each image disregarding the classes.

\subsection{Experimental Setup}\label{img_exps}

The default validation framework on the image datasets (i.e. AwA, NUS-WIDE and SUN397) consists of comparing our method against baselines and state-of-the-art methods. In addition, for all experiments the spatial feature extraction is done by applying a pre-trained CNN underneath the classification algorithm.

\subsubsection{AwA: Layered Prediction with Label Relations} \label{awa_exp}
This experiment demonstrates the label prediction capability of our SINN model and the effectiveness of adding structured label relations for label prediction. We run each method five times with five random splits – 60\% for training and 40\% for test. We report the average performance as well as the standard deviation of each performance measure. 
Note that there is very little related work with layered label prediction on AwA. The most relevant one is work by Hwang and Sigal \cite{NIPS2014_5289} on unified semantic embedding (USE). The comparison is not strictly fair, as the train/test splits are different. Further, we include our BINN model without specifying the label relation graphs (see Section \ref{binn_sec}) as a baseline method in this experiment, as it can verify the performance gain in our model from including structure.

\subsubsection{NUS-WIDE: Multi-label Classification with Partial Human Labels of Tags and Groups} \label{nus_wide_exp}
This experiment shows our model’s capability to use noisy tags and structured tag-label relations to improve multi-label classification. As previous work used various evaluation metrics and experiment settings, and there are no fixed train/test splits, it is hard to make direct comparisons. Also note that a fraction of previously used images are unavailable now due to Flickr copyright.

In order to make our result as comparable as possible, we tried to set up the experiments according to previous work. We collected all available images and discard images with missing labels as previous work did \cite{journals/corr/GongJLTI13, Johnson2015ICCV}, and got 168,240 images of the original dataset. To make our result comparable with \cite{Johnson2015ICCV}, we use 5 random splits with the same train/test ratio as \cite{Johnson2015ICCV} -- there are 132,575 training images and 35,665 test images in each split.
To compare our method with \cite{Johnson2015ICCV, conf/eccv/McAuleyL12}, we also used the tags and metadata groups in our experiment. Different from their settings, instead of augmenting images with 5000 tags, we only used 1000 tags, and augment the image with 698 group labels obtained from image medatada to form a three-layer group-concept-tag graph. Instead of using the tags as sparse binary input features (as in \cite{Johnson2015ICCV, conf/eccv/McAuleyL12}), we convert them to observed labels and feed them to our model.

We report our results on this dataset with two settings for our SINN, the first using 1k tags as the only observations to a bottom level of the relation graph. This method provides a good comparison to the tag neighborhood + tag vector [14], as we did not use extra information other than tags. In the second setting, we make both group and tag levels observable to our SINN, which achieves the best performance. We also compared our results with that of McAuley \emph{et al.} \cite{conf/eccv/McAuleyL12}, Gong \emph{et al.} \cite{journals/corr/GongJLTI13}.

\subsubsection{SUN397: Improving Scene Recognition with and without partially Observed Labels} \label{sun397_exp}
We conducted two experiments on the SUN397 dataset. The first experiment is similar to the study on AwA: we applied our model to layered image classification with label relations, and compare our model with CNN + Logistics and CNN + BINN baselines, as well as a state-of-the-art approach \cite{Xiao:2016:SDE:2963034.2963064, 5539970}. For fair comparison, we used the same train/test split ratio as \cite{Xiao:2016:SDE:2963034.2963064, 5539970}, where we have 50 training and test images in each of the 397 scene categories. To migrate the randomness in sampling, we also repeat the experiment 5 times and report the average performance as well as the standard deviations.
In the second experiment, we considered partially observed labels from the top (coarsest) scene layer as input to our inference framework. In other words, we assume we know whether an image is \textit{indoor}, \textit{outdoor man-made}, or \textit{outdoor natural}.

\subsection{Implementation Details}\label{img_details}
To optimize our learning objective, we use stochastic gradient descent with mini-batch size of 50 images and momentum of 0.9. For all training runs, we apply a two-stage policy as follows. In the first stage, we fixed pretrained CNN networks, and train our SINN with a learning rate of 0.01 with fixed-size decay step. In the second stage, we set the learning rate as 0.0001 and fine-tune the CNN together with our SINN. We set the gradient clipping threshold to be 25 to prevent gradient explosion. The weight decay value for our training procedure is set to 0.0005. In the computation of visual activations from the CNN, as different experiment datasets describe different semantic domains, we adopt different pretrained CNN models: ImageNet pretrained model \cite{caffe} for experiments \ref{awa_exp} and \ref{nus_wide_exp}, placenet pretrained model \cite{NIPS2014_5349} for experiment \ref{sun397_exp}.

\subsection{Experimental Results}\label{img_results}

For all image benchmarks, the experimental results show that the inclusion of label relation graphs effectively boost performance on the three datasets tested in \cite{inn} (i.e. AwA, SUN397 and NUS-WIDE). From Table \ref{table_awa}, \ref{table_sun397} and \ref{table_nuswide}, we can see that SINN consistently improves performance significantly in most experiments executed.

Table \ref{table_awa} shows that our method outperforms the baseline methods (CNN + Logistics and CNN + BINN variants) as well as the USE method, in terms of each concept layer and each performance metric. It validates the efficacy of our proposed model for image classification. Note that for the results in Table 1, we did not finetune the first seven layers of the CNN \cite{Krizhevsky:2017:ICD:3098997.3065386} for fairer comparison with Hwang and Sigal \cite{hwang2012semantic} (which only makes use of DECAF features \cite{pmlr-v32-donahue14}). Fine-tuning the first seven CNN layers further improves IoU\textsubscript{Acc} at each concept layer to 86.06 \textpm 0.72 (28 taxonomy terms), 69.17 \textpm 1.00 (50 animal classes), 88.22 \textpm 0.38 (85 attributes), and mAP\textsubscript{L} to 94.17 \textpm 0.55 (28 taxonomy terms), 83.12 \textpm 0.69 (50 animal classes), 96.72 \textpm 0.20 (85 attributes), respectively.

The results on SUN397 are summarized in Table \ref{table_sun397}, showing that our proposed method again achieves a considerable performance gain over all the compared methods. In Table \ref{table_sun_partial}, we compare the 397 fine-grained scene recognition performance. We compare to a set of baselines, including CNN + Logistics + Partial Labels that considers the partial labels as an extra binary indicator feature vector for logistic regression. Results show that our method combined with partial labels (i.e., CNN + SINN + Partial Labels) improves over baselines, exceeding the second best by 4\% MC\textsubscript{Acc} and 6\% mAP\textsubscript{L}.

The results on NUS-WIDE are shown in Table \ref{table_nuswide}. We can see that SINN outperformed all the baseline methods and existing approaches (i.e.\cite{Johnson2015ICCV, journals/corr/GongJLTI13, conf/eccv/McAuleyL12}) by a large margin on NUS-WIDE dataset, considering different settings. The results on SUN397 are summarized in Table \ref{table_nuswide} and present consistent improvements when using SINN.

\begin{table*}[!t]
\centering
\begin{tabular}{|l||c|c|c|c|}
    \hline Concept Layer & Model & MC Acc & IoU & $mAP_L$ \\
    \Xhline{4\arrayrulewidth} \multirow{3}{*}{28 taxonomy terms} & CNN + Logistics & - & 80.41 \textpm 0.09 & 90.16 \textpm 0.10 \\
                                                                 & CNN + BINN & - & 79.85 \textpm 0.13 & 89.92 \textpm 0.07 \\
                                                                 & CNN + SINN & - & \textbf{84.47 \textpm 0.38} & \textbf{93.00 \textpm 0.29} \\

    \Xhline{4\arrayrulewidth} \multirow{4}{*}{50 animal classes} & USE \cite{NIPS2014_5289} + DECAF \cite{pmlr-v32-donahue14} & 46.42 \textpm 1.33 & - & - \\
                                                                 & CNN + Logistics & 78.44 \textpm 0.27 & 62.75 \textpm 0.26 & 78.35 \textpm 0.19 \\
                                                                 & CNN + BINN & 79.00 \textpm 0.43 & 62.80 \textpm 0.25 & 78.88 \textpm 0.35 \\
                                                                 & CNN + SINN & \textbf{79.36 \textpm 0.43} & \textbf{66.60 \textpm 0.43} & \textbf{81.19 \textpm 0.14} \\
                                                                 
    \Xhline{4\arrayrulewidth} \multirow{3}{*}{85 attributes} & CNN + Logistics & - & 81.29 \textpm 0.10 & 93.29 \textpm 0.12 \\
                                                                 & CNN + BINN & - & 80.64 \textpm 0.13 & 93.04 \textpm 0.13 \\
                                                                 & CNN + SINN & - & \textbf{86.92 \textpm 0.18} & \textbf{96.05 \textpm 0.07} \\
    \hline
\end{tabular}
\caption{Layered label prediction results on the AwA dataset.}
\label{table_awa}
\end{table*}

\begin{table*}[!t]
\centering
\begin{tabular}{|l||c|c|c|c|}
    \hline Concept Layer & Model & MC Acc & IoU & $mAP_L$ \\
    \Xhline{4\arrayrulewidth} \multirow{3}{*}{3 coarse scene categories} & CNN + Logistics & - & 83.67 \textpm 0.18 & 95.19 \textpm 0.07 \\
                                                                 & CNN + BINN & - & 83.63 \textpm 0.24 & 95.19 \textpm 0.03 \\
                                                                 & CNN + SINN & - & \textbf{85.95 \textpm 0.44} & \textbf{96.40 \textpm 0.18} \\

    \Xhline{4\arrayrulewidth} \multirow{4}{*}{16 general scene categories} & CNN + Logistics & - & 64.30 \textpm 0.27 & 83.30 \textpm 0.19\\
                                                                 & CNN + BINN & - & 63.40 \textpm 0.35 & 82.93 \textpm 0.14 \\
                                                                 & CNN + SINN & - & \textbf{66.46 \textpm 1.10} & \textbf{84.97 \textpm 0.96} \\
                                                                 
    \Xhline{4\arrayrulewidth} \multirow{3}{*}{397 fine-grained scene categories} & Image features + SVM (\cite{Xiao:2016:SDE:2963034.2963064, 5539970}) & 42.70 & - & - \\
                                                                 & CNN + Logistics & \textbf{57.86 \textpm 0.18} & 35.97 \textpm 0.37 & 55.31 \textpm 0.30\\
                                                                 & CNN + BINN & 57.52 \textpm 0.29 & 35.44 \textpm 1.02 & 55.57 \textpm 0.63\\
                                                                 & CNN + SINN & 57.60 \textpm 0.38 & \textbf{37.71 \textpm 1.13} & \textbf{58.00 \textpm 0.33} \\

    \hline
\end{tabular}
\caption{Layered label prediction results on the SUN397 dataset.}
\label{table_sun397}
\end{table*}

\begin{table}[!t]
\centering
\begin{tabular}{|l||c|c|c|c|}
    \hline Model & MC+Acc & mAP$_L$ \\
    \hline Image features + SVM \cite{Xiao:2016:SDE:2963034.2963064, 5539970} & 42.70 & - \\
    \hline CNN + Logistics & 57.86 \textpm 0.38 & 55.31 \textpm 0.30 \\
    \hline CNN + BINN & 57.52 \textpm 0.29 & 55.57 \textpm 0.63 \\
    \hline CNN + SINN & \textbf{57.60 \textpm 0.38} & \textbf{58.00 \textpm 0.33} \\
    \Xhline {4\arrayrulewidth} CNN + Logistics + Partial Labels & 59.08 \textpm 0.27 &  56.88 \textpm 0.29\\
    \hline CNN + SINN + Partial Labels & \textbf{63.46 \textpm 0.18} & \textbf{64.63 \textpm 0.28} \\

    \hline
\end{tabular}
\caption{Recognition results on the 397 fine-grained scene categories. Note that the last two compared methods make use of partially observed labels from the top (coarsest) scene layer, i.e. \textit{indoor}, \textit{outdoor man-made}, and \textit{outdoor natural}.}
    \label{table_sun_partial}
\end{table}

\begin{table*}[!t]
\centering
\begin{tabular}{|l||c|c?c|c|c|c|}
    \hline Model & mAP\textsubscript{L} & mAP\textsubscript{I} & R\textsubscript{L} & P\textsubscript{L} & R\textsubscript{I} & R\textsubscript{I}  \\
    \Xhline{4\arrayrulewidth} Graphical Model \cite{conf/eccv/McAuleyL12} & 49.00 & - & - & - & - & - \\
    \hline CNN + WARP \cite{journals/corr/GongJLTI13} & - & - & 35.60 & 31.65 & 60.49 & 48.59 \\
    \hline 5k tags + Logistics \cite{Johnson2015ICCV}  & 43.88 \textpm 0.32 & 77.06 \textpm 0.14 & 47.52 \textpm 2.59 & 46.83 \textpm 0.89 & 71.34 \textpm 0.16 & 51.18 \textpm 0.16 \\
    \hline Tag neighbors + 5k tags \cite{Johnson2015ICCV}  & 61.88 \textpm 0.36 & 80.27 \textpm 0.08 & 57.30 \textpm 0.44 & 54.74 \textpm 0.63 & 75.10 \textpm 0.20 & 53.46 \textpm 0.09 \\
    \Xhline{4\arrayrulewidth} CNN + Logistics & 46.94 \textpm 0.47 & 72.25 \textpm 0.19 & 45.03 \textpm 0.44 & 45.60 \textpm 0.35 & 70.77 \textpm 0.21 & 51.32 \textpm 0.14 \\
    \hline 1k tags + Logistics & 50.33 \textpm 0.37 & 66.57 \textpm 0.12 & 23.97 \textpm 0.23 & 47.40 \textpm 0.07 & 64.95 \textpm 0.18 & 47.40 \textpm 0.07 \\
    \hline 1k tags + Groups + Logistics & 52.81 \textpm 0.40 & 68.04 \textpm 0.12 & 25.54 \textpm 0.24 & 49.26 \textpm 0.15 & 65.99 \textpm 0.15 & 48.13 \textpm 0.05 \\
    \hline 1k tags + Groups + CNN + Logistics & 54.67 \textpm 0.57 & 77.81 \textpm 0.22 & 50.83 \textpm 0.53 & 49.36 \textpm 0.30 & 75.38 \textpm 0.16 & 54.61 \textpm 0.09 \\
    \Xhline{4\arrayrulewidth} 1k tags + CNN + SINN & 67.20 \textpm 0.60 & 81.99 \textpm 0.14 & 59.82 \textpm 0.12 & 57.02 \textpm 0.57 & 78.78 \textpm 0.13 & 56.84 \textpm 0.07 \\
    \hline 1k tags + Groups + CNN + SINN & \textbf{69.24 \textpm 0.47} & \textbf{82.53 \textpm 0.15} & \textbf{60.63 \textpm 0.67} & \textbf{58.30 \textpm 0.33} & \textbf{79.12 \textpm 0.18} & \textbf{57.05 \textpm 0.09} \\
    \hline
    
\end{tabular}
\caption{Results on NUS-WIDE for  from \cite{inn}. We also included the results for per-label recall and precision (R\textsubscript{L} and P\textsubscript{L}) and per-image (R\textsubscript{I} and P\textsubscript{I}), considering the top 3 labels for each image.} 
    \label{table_nuswide}
\end{table*}

\section{Multi-label Classification on Videos} \label{vid_task}
In this section, we describe the work done for multi-label classification on videos. In Section \ref{vid_datasets} and \ref{vid_metrics}, we present details on the two YouTube-8M benchmarks as well as the metrics used for this task, respectively. Section \ref{vid_exps} and \ref{vid_details}, present our experimental setup and implementation details, in this order. Lastly, we report the results obtained in Section \ref{vid_results}. 

\subsection{Evaluation Benchmarks}\label{vid_datasets}
\subsubsection{YouTube-8M (YT-8M)}
The YouTube-8M dataset consists of approximately 8 million YouTube videos,
each annotated with 4800 Google Knowledge Graph \textit{entities}, functioning as
classes. With each entity label is associated up to 3 \textit{verticals} (i.e.\ coarse-grained labels). The dataset is derived from roughly 500K hours of
untrimmed videos, with an average of 1.8 labels per video. Each video is decoded
as a set of features extracted by passing the RGB frame through the InceptionV3 model from Szegedy \emph{et al.} \cite{inceptionv3},
a deep CNN pretrained on ImageNet \cite{imagenet}, and Principal Component Analysis (PCA) 
is applied to reduce feature dimension. The scale of this dataset in both label 
space and data space is unprecedented in the field of video datasets, surpassing 
previous benchmarks such as UCF-101 and Sports1M.

\subsubsection{YouTube-8M V2 (YT-8M V2)}
The YouTube-8M V2 dataset represents the frame and audio features from approximately 7 million YouTube videos. The dataset is an updated version of YouTube-8M, with an increased number of labels per video and a smaller number of entities. On average, the videos in YT-8M V2 have 3.4 labels each, and there are only 4716 Google Knowledge Graph entities forming the label space. The preprocessing for this dataset is the same as YT-8M, but the audio features are also included, calculated using the CNN method in \cite{audio}.

\subsection{Evaluation Metrics}\label{vid_metrics}
The metrics formulated in this section considers a collection of videos $\mathcal{V}$, such that for each video $v \in \mathcal{V}$, a model predicts a set of labels $\mathcal{P}_v \in \mathcal{C}$ and the metric is computed against a set of ground-truth labels $\mathcal{G}_v \in \mathcal{C}$. We start by defining the indicator function for a generic set $\mathcal{A}$ and an element $x$, used in further formulations, as follows:
\begin{equation}
\mathds{1}_\mathcal{A}(x)=
\begin{cases}
  1, & \text{if}\ x \in \mathcal{A}, \\
  0, & \text{otherwise}.
\end{cases}
\end{equation}



\subsubsection{Hit at Top K (Hit@k)} \label{hit@k}
The $\it{Hit@k}$ corresponds to the portion of samples whose top $k$ predictions contain at least one ground-truth label and can be written as follows:
\begin{equation}
\it{Hit@k(\mathcal{V})} =  \frac{1}{|\mathcal{V}|}\sum_{v \in \mathcal{V}} \vee_{e \in \it{top}(\mathcal{P}_v, k)} \mathds{1}_{\mathcal{G}_v}(e),
\end{equation}
where $\vee$ is the logical OR operator.

\subsubsection{Precision at Equal Recall Rate (PERR)} \label{PERR}
The definition for this metric is the same as in \cite{youtube8m} and consists of the precision computed for each sample when retrieving the same number of labels as the ground-truth labels: 
\begin{equation}
\it{PERR(\mathcal{V})} =  \frac{1}{|\mathcal{V}'|}\sum_{v \in \mathcal{V}'} \Bigg[ \frac{1}{|\mathcal{G}_v|} \sum_{e \in \it{top}(\mathcal{P}_v, |\mathcal{G}_v|)} \mathds{1}_{\mathcal{G}_v}(e) \Bigg],
\end{equation}
where $\mathcal{V}' \subseteq \mathcal{V}$ is the subset of test videos containing at least one ground-truth label.

\subsubsection{Global Average Precision (gAP)}  \label{gap}
The gAP metric defined in \cite{youtube8m} is computed similarly to the mAP\textsubscript{L}(Section \ref{lmap}) defined earlier, with the distinction that the gAP is computed agnostically to the top k (i.e. k=20) classes as for a binary problem:

\begin{equation}\label{eqn:gap_equation}
\it{gAP} =  \sum_{i =1}^{kN_v} P(i) \Delta R(i),
\end{equation}
where $N_v$ is the number of videos in the test set, P(i) and $\Delta R(i)$ are defined as for Eq. \eqref{eqn:ap_equation} 

\subsubsection{Per Video Mean Average Precision (mAP\textsubscript{V})}  \label{vmap}
As is \cite{youtube8m}, buckets of length $10^{-4}$ are used for discretizing the precision-recall threshold $\tau$ for each class. All the non-zero annotations are then sorted (ascending order) according to the prediction scores. The precision $P_c(\tau)$ and recall $R_c(\tau)$ at a given threshold $\tau$ for class $c$ is defined as follows:
\begin{equation}\label{eqn:precision} 
\it{P_c}(\tau) =  \frac{\sum_{i = 1}^{N_v} (y_{ic} \geq \tau)t_{ic}}{\sum_{i = 1}^{N_v} (y_{ic} \geq \tau)},
\end{equation}

\begin{equation}\label{eqn:recall}
\it{R_c}(\tau) =  \frac{\sum_{i = 1}^{N_v} (y_{ic} \geq \tau)t_{ic}}{\sum_{i = 1}^{N_v} y_{ic}},
\end{equation}
where $y_{ic}$ and $t_{ic}$ are prediction and ground-truth for the i-th video for class $c$ and $N_v$ is the total number of videos.

The $AP_c$ for a specific class $c$ is given by:
\begin{equation}\label{eqn:ap_youtube8m}
\it{AP_c} =  \sum_{j = 1}^{10^{4}} P_c(\tau_j)[R_c(\tau_j)-R_c(\tau_{j+1})],
\end{equation}
where $\tau_j = \frac{j}{10^4}$. The mAP\textsubscript{V} is obtained by taking the unweighted mean across all classes in $\mathcal{C}$ (as in Eq. \eqref{eqn:ap_avgl}).

\subsection{Experimental Setup} \label{vid_exps}
In order to validate our method on multi-label video classification, we follow a similar framework to the one used for multi-label image classification (Section \ref{img_exps}). Firstly, we reproduce the logistic baseline from \cite{youtube8m} and compare against our models for both releases of YouTube-8M. For YT-8M V2, we conduct experiments with spatial features only and spatial plus audio features, comparing our method against the logistic baseline. Additionally, following the same intuition from the partially observed experiments for multi-label image classification. We run experiments providing the partial observations of the verticals for both datasets to our model and validate against the logistic baseline including the partial observations as input features. Note that including partial observations to the logistic regression is important to demonstrate that partial labels are not the discriminative feature for these experiments. All experiments were executed following the guidelines presented in \cite{youtube8m} and the official Kaggle competition. 

\subsection{Implementation Details} \label{vid_details}
The dataset labels were organized into a graph with two concept layers -- entities (i.e. fine-grained labels), and verticals (i.e. coarse-grained labels). We minimize the cross-entropy loss function using the Adam optimizer \cite{adam}. For all models, mini-batches of size 1024 were used, and a weight decay of $10^{-8}$ was applied. The logistic regression model was trained for 35k iterations with a learning rate of 0.01, and the BINN was trained for 90k iterations, starting with a learning rate of 0.001 with a decay factor of 0.1 at every 40k iterations. All models were implemented with the Caffe deep learning framework \cite{caffe}.

\subsection{Experimental Results} \label{vid_results}
In Table \ref{table1}, results on YT-8M are presented on the validation set, where the first group of results refers to the baseline models reported in \cite{youtube8m}. To ensure consistency with \cite{youtube8m}, the results for the logistic regression baseline trained by us is shown in the second group of models in Table \ref{table1}. It is worth to mention that the inclusion of Z- and L2-normalization were crucial for duplicating the baseline results and boosting results for the other models. As can be verified from Table \ref{table1}, the best SINN model for YT-8M achieved significant improvements on all metrics. More precisely, a 3.2\% improvement was seen on mAP using exclusively video-level features, against the leading baseline. The SINN also demonstrated measurable success on the PERR, Hit@1 and gAP metrics, with respective improvements of 4.69\%, 4.4\% and 7.41\%.

The results shown in Table \ref{table2} follow the same sequence as the previous results, save for the inclusion of audio features available for YT-8M V2. The best results were obtained using RGB and audio features, with Z- and L2-Normalization. The most effective SINN model obtained 42.32\%, 72.92\%, 85.49\% and 80.09\% for mAP, PERR, Hit@1 and gAP respectively, which corresponds to improvements of 3.71\%, 3.64\%, 2.74\% and 4.26\%, against the leading baseline results. Additionally, our results indicate that our method for combining partial labels also achieves significant improvements over baselines on both releases of YouTube-8M. 

\begin{table}[!t]
\centering
\begin{tabular}{|l||c|c|c|c|}
    \hline Model & mAP\textsubscript{V} & PERR & Hit@1 & gAP \\
    \Xhline{4\arrayrulewidth} LSTM \cite{youtube8m} & 26.60 & 57.30 & 64.50 & - \\
    \hline Mixture-of-Experts \cite{youtube8m} & 30.00 & 55.80 & 63.30 & -\\
    \hline Logistics \cite{youtube8m} & 28.10 & 53.00 & 60.50 & -\\
    \Xhline {4\arrayrulewidth} Logistics \cite{DBLP:journals/corr/NauataSM17} &  27.98 & 52.89& 60.34 & 49.04\\
    \hline BINN \cite{DBLP:journals/corr/NauataSM17} &30.17 & 57.40 & 64.48 &55.76 \\
    \hline SINN  & \textbf{31.18} & \textbf{57.58} & \textbf{64.74} & \textbf{56.39}\\
    \Xhline {4\arrayrulewidth} Logistics + Partial labels & 54.47 & 68.77 & 75.53 & 71.46 \\
    \hline SINN + Partial labels & \textbf{57.71} & \textbf{72.82} & \textbf{78.85} & \textbf{75.80} \\
    \hline
\end{tabular}
\caption{YouTube-8M (YT-8M) results for mAP\textsubscript{V}, Precision at Equal Recall Rate (PERR), Hit at 1 and gAP on the validation set.}
\label{table1}
\end{table}

\begin{table}[!t]
\centering
\begin{tabular}{|l||c|c|c|c|}
    \hline Model & mAP\textsubscript{V} & PERR & Hit@1 & gAP \\
    \hline Logistics + RGB  \cite{DBLP:journals/corr/NauataSM17} & 36.84 & 64.38 & 78.62 & 70.31 \\
    \hline Logistics + RGB + Audio \cite{DBLP:journals/corr/NauataSM17} & 38.61 & 69.28 & 82.75 & 75.83\\
    \Xhline {4\arrayrulewidth} BINN + RGB \cite{DBLP:journals/corr/NauataSM17} & 39.12 &  69.01 & 82.27 & 75.99\\
    \hline SINN + RGB   & 40.19 & 69.11 & 82.35 & 76.33\\
    \hline BINN + RGB + Audio \cite{DBLP:journals/corr/NauataSM17}  & 40.91 & 72.27 & 84.96 & 79.29\\
    \hline SINN + RGB + Audio  &  \textbf{42.32} & \textbf{72.92} & \textbf{85.49} & \textbf{80.09}\\
    \Xhline {4\arrayrulewidth} Logistics + Partial labels & 50.28 & 78.52 & 89.82 & 84.94 \\
    \hline SINN + Partial labels & \textbf{51.88} & \textbf{80.55} & \textbf{90.99} & \textbf{86.51} \\
    \hline
\end{tabular}
\caption{YouTube-8M V2 (YT-8M V2) results for mAP\textsubscript{V}, Precision at Equal Recall Rate (PERR), Hit at 1 and gAP on the validation set.}
    \label{table2}
\end{table}

\section{Dense Action Detection} \label{det_task}
Here we introduce the work done on dense action detection. Details about THUMOS'14 and MultiTHUMOS benchmarks are presented in Section \ref{det_datasets} and metrics used for this task in Section \ref{det_metrics}. Our experimental setup is discussed in Section \ref{det_exps} and implementation details in Section \ref{det_details}. The results are reported in Section \ref{det_results}.

\subsection{Evaluation Benchmarks}\label{det_datasets}
\subsubsection{THUMOS'14}
The THUMOS'14  \cite{THUMOS14} dataset contains 13,320 single-labeled trimmed videos from UCF101 \cite{ucf101} and an extra 2,584 multi-labeled untrimmed videos, from which 1,010 belong to the validation set and 1,574 to the test set. In addition to the entire set of trimmed videos, around 400 untrimmed videos (out of 2,584) are utilized for the temporal action detection task. On average these videos (i.e.\ trimmed and untrimmed) used for action detection, contain approximately a single label per video and 20 different action classes.

\subsubsection{MultiTHUMOS}
In order to study the problem of densely annotated unconstrained videos, Yeung \emph{et al.} introduced the MultiTHUMOS dataset \cite{yeung2015every}. This dataset consists of an augmented version of the untrimmed videos for action detection from THUMOS'14, corresponding to around 30 hours of temporal action annotation. More precisely, the MultiTHUMOS dataset extends the annotation from 6,365 over 20 action classes in THUMOS'14 to 38,690 annotations over 65 action classes, including 32,325 annotations over the 45 introduced action classes. In addition, the density of annotations increases from 0.3 to 1.5 action classes per frame in average and from 1.1 to 10.5 action classes per video.

\subsection{Evaluation Metrics} \label{det_metrics}

Analogously to Section \ref{lmap}, we use mAP\textsubscript{L} to evaluate our method, with the difference that for this task each sample corresponds to one single frame and the mAP\textsubscript{L} is computed across all videos as in a long global list of predictions for a binary problem. As in \cite{yeung2015every}, this metric is utilized in order to avoid aggregating frame-level predictions for obtaining activities segments.

\subsection{Experimental Setup} \label{det_exps}
As a standard approach for this dataset, we train our model on the untrimmed validation set and validate on the test set, since the training set is the same from UCF101 and consists of short clipped videos. We executed experiments to validate the inclusion of structured annotation in two scenarios: our inference modules versus CNN in static frames and structured temporal inference versus LSTM baseline, allowing message propagation across time.

Firstly, we reproduce the single-frame CNN baseline presented in \cite{yeung2015every}, to certify that we are extracting competitive features. This baseline is compared against our single-frame models, i.e. BINN and SINN. Next, we train a LSTM baseline to compare against our temporal extention models, i.e. biLSTM and siLSTM. As in \cite{yeung2015every}, we report mAP\textsubscript{L} scores for MultiTHUMOS and the subset of classes from THUMOS on the untrimmed test set from MultiTHUMOS.

\subsection{Implementation Details} \label{det_details}
The single-frame CNN baseline was pretrained on Imagenet and then fine-tuned on MultiTHUMOS. We extracted 4096-dimensional features from individual frames at fc-7 in the base model similar procedure done in \cite{yeung2015every}. 

In order to obtain structured annotations for our models, we augment the original set of labels $\mathcal{L}$ by manually converting each $l \in \mathcal{L}$ to an approximate word sense from WordNet, generating a new set $\mathcal{L}'$. We select new concepts for each $l' \in \mathcal{L}'$ that surpass a certain \textit{relatedness score} (e.g. vector score \cite{Patwardhan2003IncorporatingDA}) threshold (e.g. 0.5) against all synsets from WordNet. The augmented set of concepts and its structure is then obtained by joining all the new concepts together with the elements from $\mathcal{L}'$ and computing pair-wise \textit{relatedness score} for all elements within the new superset. Parsing WordNet provided us an extra layer containing 32 new concepts whose connections to the original MultiTHUMOS labels were determined by the \textit{relatedness score}. The single-frame BINN and SINN models were trained with a batch size of 1024 frames for 8500 iterations, step size of 5000 iterations, learning rate of $10^{-3}$ and decay of 0.1.

For the LSTM baseline, we use the same feature vectors extracted for the single-frame CNN baseline. Differently from \cite{yeung2015every}, our LSTM based models are fed with randomly sampled contiguous sequences of 32 frames (~3.2s) from each video sample. In particular, the LSTM baseline was trained using ADAM with a learning rate of $10^{-3}$, step size of 1500 iterations, batch size of 512 and learning rate decay of 0.1 for 1500 iterations.

Our proposed models biLSTM and siLSTM were designed to include temporal dependencies as LSTM plus, hierarchically structured labels. For these models we use the same 32 extra concepts extracted from WordNet. Also, we use a similar approach as for the LSTM model for feeding frame sequences to these models. The best model (i.e.\ siLSTM) obtained was trained using ADAM, batch size of 1024, learning rate of $7\times10^{-4}$ for 900 iterations, decaying 0.1 every 700 iterations.

\subsection{Experimental Results} \label{det_results}
\begin{figure*}
  \begin{subfigure}[t]{.5\textwidth}
    \centering
    \includegraphics[width=\linewidth]{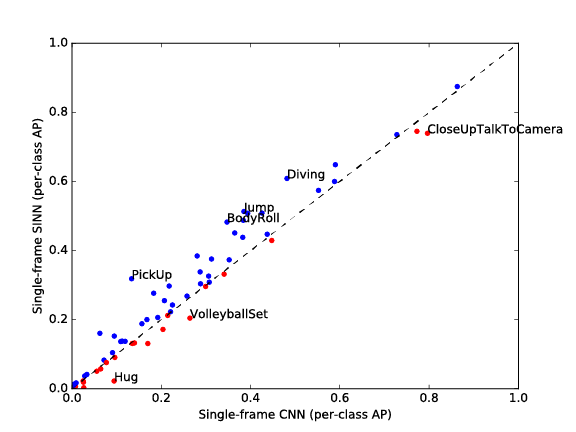}
    \caption{Per-class APs for Single-frame SINN versus Single-frame CNN (VGG).}
    \label{fig:binn_vs_cnn}
  \end{subfigure}
  \hfill
  \begin{subfigure}[t]{.5\textwidth}
    \centering
    \includegraphics[width=\linewidth]{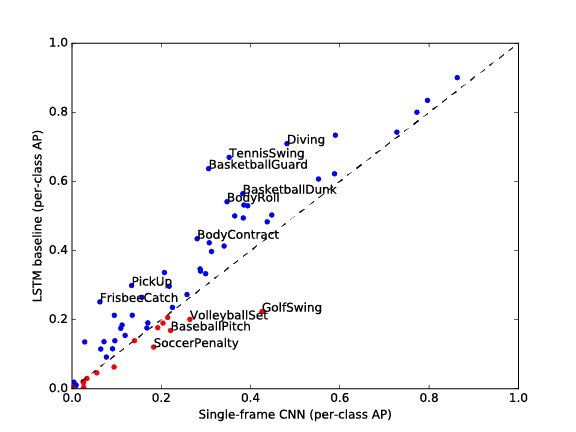}
    \caption{Per-class AP for LSTM versus Single-frame CNN (VGG).}
    \label{fig:lstm_vs_cnn}
  \end{subfigure}

  \medskip

  \begin{subfigure}[t]{.5\textwidth}
    \centering
    \includegraphics[width=\linewidth]{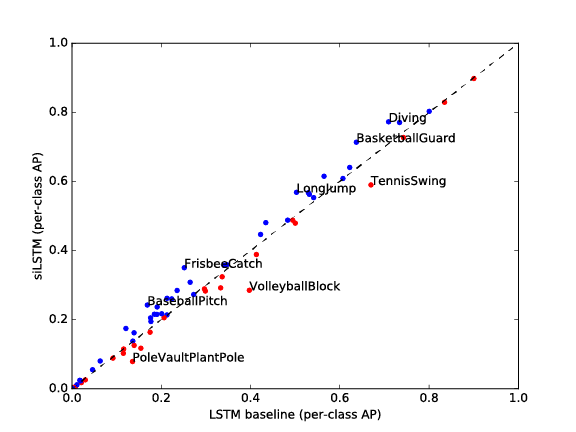}
    \caption{Per-class AP for siLSTM versus LSTM.}
    \label{fig:ilstm_vs_lstm}
  \end{subfigure}
  \hfill
  \begin{subfigure}[t]{.5\textwidth} 
    \centering
    \includegraphics[width=\linewidth]{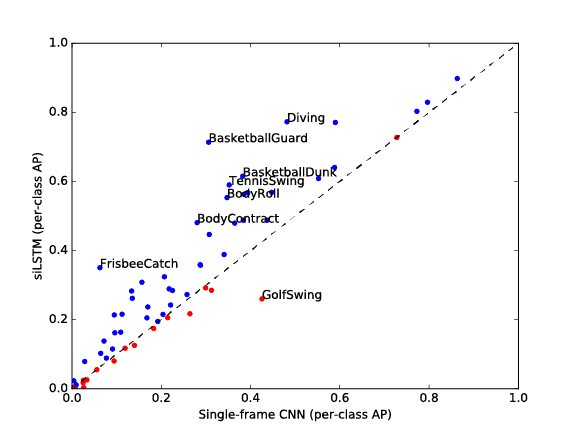}
    \caption{Per-class AP for siLSTM versus Single-frame CNN (VGG).}
    \label{fig:ilstm_vs_cnn}
  \end{subfigure}
\end{figure*}

As shown in Table \ref{table3}, the inclusion of structured labels parsing WordNet using the single-frame SINN provides a gain of 2.7\% on THUMOS and ~2.4\% on MultiTHUMOS against the single-frame CNN baseline implemented by us. In Figure \ref{fig:binn_vs_cnn}, we show a comparison between the single-frame SINN and the single-frame CNN using per-class AP computer for each model. 

As observed in \cite{yeung2015every}, the inclusion of temporal dependencies clearly provides substantial improvements in activity detection as we can also conclude from results shown in Table \ref{table3}. The LSTM model implemented by us provides 6.3\% THUMOS and 5.9\% on MultiTHUMOS over our single-frame CNN baseline. The APs breakdown is shown in Figure \ref{fig:lstm_vs_cnn}. 

The results from our proposed models (i.e. biLSTM and siLSTM) show some gain in performance over the LSTM baseline as show in Table \ref{table3}. More precisely, biLSTM and siLSTM are benefiting from both inclusion of structured labels and temporal label dependencies. According to our experiments, the siLSTM provides a boost of 2.1\% on THUMOS and 1.0\% on MultiTHUMOS over the LSTM baseline as we can see from Table \ref{table3}. In Figure \ref{fig:ilstm_vs_lstm}, we present class-specific APs comparison between the siLSTM and LSTM. In Figure \ref{fig:ilstm_vs_cnn}, we present the detailed overall improvement achieved, using siLSTM over the single-frame CNN baseline.

The significant difference obtained between our LSTM baseline and the LSTM results from \cite{yeung2015every} impede a direct and fair comparison with MultiLSTM. However, analyzing the relative gain between siLSTM and MultiLSTM over the corresponding baselines indicate that siLSTM is competitive with MultiLSTM. As presented before the relative gain provided by siLSTM is 2.1\% and 1.0\% against 2.0\% and 1.6\% for MultiLSTM on THUMOS and MultiTHUMOS, respectively. It is worth to mention that siLSTM and MultiLSTM explore different ideas for improving over LSTM, which are not mutually exclusive thus, they can possibly be combined for obtaining a potentially stronger model.

\begin{figure*}

\includegraphics[width=1.0\linewidth]{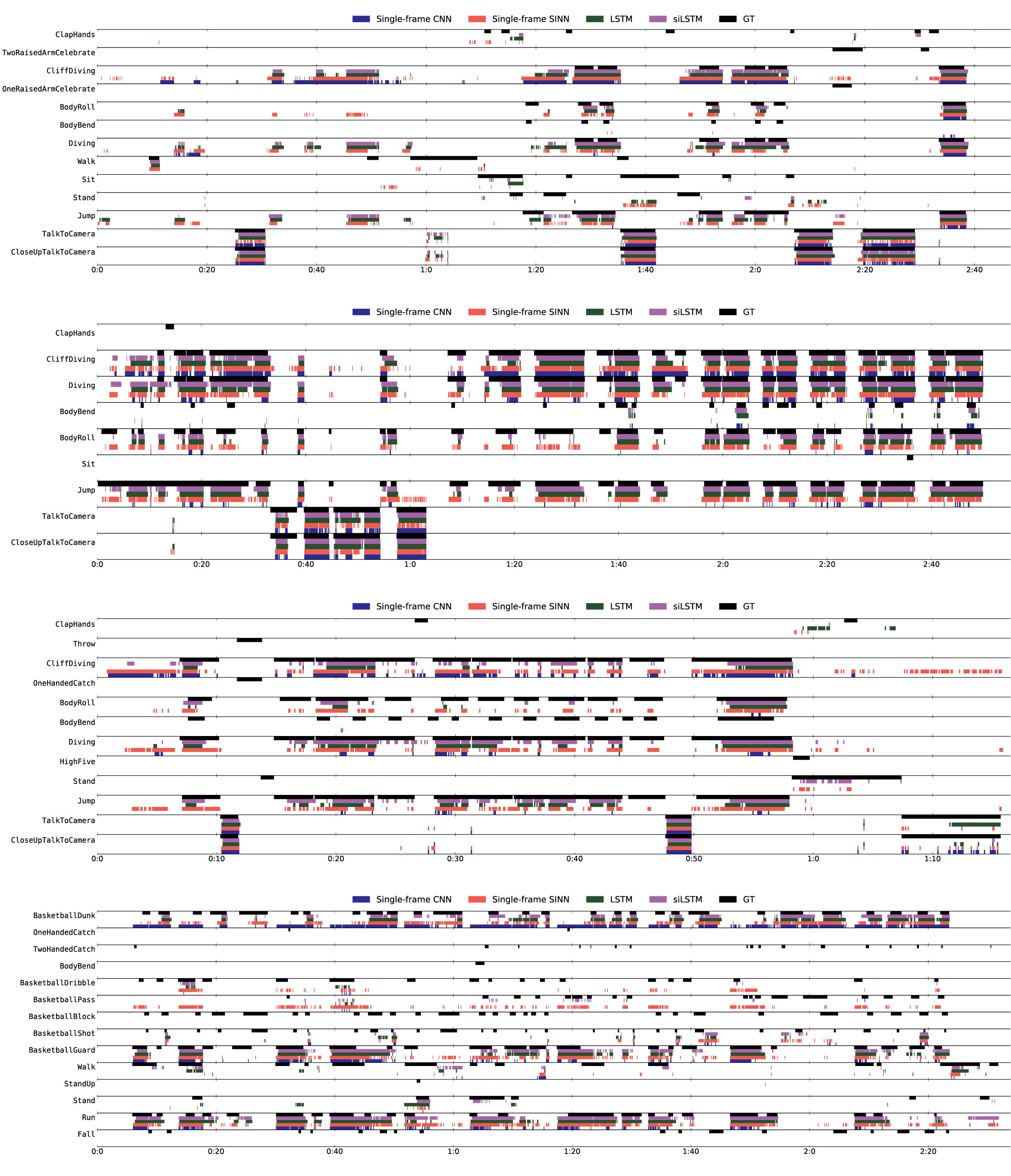}
\caption{Timeline comparison for single-frame models (CNN and SINN) and sequential models (LSTM and siLSTM) in multi-label action detecion for four different videos. The verical axis represents the ground-truth labels appearing at least once in the target video and the horizontal axis corresponds to the duration of the target video in minutes. (Best view in color)}
\label{fig:timeline_comparison}
\end{figure*}

Figure \ref{fig:timeline_comparison} shows a timeline comparison for four different videos in their entirety, comparing a set of key models covered in this work (i.e. single-frame CNN and SINN, LSTM, siLSTM). The series points for prediction in this plot were obtained by thresholding the prediction at 0.5, similar procedure was done in \cite{yeung2015every}.

\begin{table}[!t]
\centering
\begin{tabular}{|l||c|c|}
    \hline Model & THUMOS & MultiTHUMOS\\
    \hline Single-frame CNN \cite{yeung2015every} & 34.7 & 25.4\\
    \hline Single-frame CNN [Ours] & 35.5 & 24.6\\
    \hline Single-frame BINN & 37.6 & 26.8\\
    \hline Single-frame SINN & \textbf{38.2} & \textbf{27.0}\\
    \Xhline {4\arrayrulewidth} LSTM \cite{yeung2015every} & 39.3 & 28.1\\
    \hline LSTM [Ours] & 41.8 & 30.5\\
    \hline MultiLSTM \cite{yeung2015every} & 41.3 & 29.7\\
    \hline biLSTM & 42.8 &  30.6\\
    \hline siLSTM & \textbf{43.9} & \textbf{31.5}\\
    \hline
\end{tabular}
\caption{THUMOS and MultiTHUMOS results for mAP\textsubscript{L}.}
    \label{table3}
\end{table}

\section{Conclusion} \label{conclusion}
In this paper we bind together previous successes in performing inference in structured graphs (i.e.\ BINN and SINN), present new results for partially observed prediction on both releases of YouTube-8M and propose an extension for propagating concepts through time, based on the LSTM formulation. We consistently show significant improvements of performance in multi-label image classification, multi-label video classification and action detection tasks across a number of public datasets, achieving considerable gains against baselines and existing approaches. 

From our experiments, the structured inference performed by BINN and SINN can lead to impressive boosts in accuracy. The results obtained for multi-label image and video classification are strongly positive and even without explicitly defining the graph structure as prior knowledge in the SINN model we were able to obtain substantial improvements on the YouTube-8M benchmarks. Additionally, the results presented using partially observed labels indicate that our method processes partial observations more effectively than the baselines and is able to predict missing labels with higher confidence.

The proposed action detection method also provides some improvements however, we believe that the lack of a more atomic structure of actions, person-centric spatial annotations and higher confidence structured annotations pose challenges for the learning of less noisy relations in label space.

%

\ifCLASSOPTIONcaptionsoff
  \newpage
\fi
\bibliographystyle{IEEEtran}
\bibliography{IEEEabrv,egbib}

\begin{thebibliography}{10}
\providecommand{\url}[1]{#1}
\csname url@samestyle\endcsname
\providecommand{\newblock}{\relax}
\providecommand{\bibinfo}[2]{#2}
\providecommand{\BIBentrySTDinterwordspacing}{\spaceskip=0pt\relax}
\providecommand{\BIBentryALTinterwordstretchfactor}{4}
\providecommand{\BIBentryALTinterwordspacing}{\spaceskip=\fontdimen2\font plus
\BIBentryALTinterwordstretchfactor\fontdimen3\font minus
  \fontdimen4\font\relax}
\providecommand{\BIBforeignlanguage}[2]{{%
\expandafter\ifx\csname l@#1\endcsname\relax
\typeout{** WARNING: IEEEtran.bst: No hyphenation pattern has been}%
\typeout{** loaded for the language `#1'. Using the pattern for}%
\typeout{** the default language instead.}%
\else
\language=\csname l@#1\endcsname
\fi
#2}}
\providecommand{\BIBdecl}{\relax}
\BIBdecl

\bibitem{imagenet}
O.~Russakovsky, J.~Deng, H.~Su, J.~Krause, S.~Satheesh, S.~Ma, Z.~Huang,
  A.~Karpathy, A.~Khosla, M.~Bernstein, A.~C. Berg, and L.~Fei-Fei, ``Imagenet
  large scale visual recognition challenge,'' \emph{International Journal of
  Computer Vision (IJCV)}, vol. 115, no.~3, pp. 211--252, 2015.

\bibitem{Chua:2009:NRW}
T.-S. Chua, J.~Tang, R.~Hong, H.~Li, Z.~Luo, and Y.~Zheng, ``Nus-wide: A
  real-world web image database from national university of singapore,'' in
  \emph{CIVR}, 2009.

\bibitem{6571196}
C.~H. Lampert, H.~Nickisch, and S.~Harmeling, ``Attribute-based classification
  for zero-shot visual object categorization,'' \emph{IEEE Transactions on
  Pattern Analysis and Machine Intelligence (TPAMI)}, vol.~36, no.~3, pp.
  453--465, 2014.

\bibitem{5539970}
J.~Xiao, J.~Hays, K.~A. Ehinger, A.~Oliva, and A.~Torralba, ``Sun database:
  Large-scale scene recognition from abbey to zoo,'' in \emph{CVPR}, 2010.

\bibitem{youtube8m}
S.~Abu{-}El{-}Haija, N.~Kothari, J.~Lee, P.~Natsev, G.~Toderici,
  B.~Varadarajan, and S.~Vijayanarasimhan, ``Youtube-8m: {A} large-scale video
  classification benchmark,'' \emph{CoRR}, 2016.

\bibitem{THUMOS14}
Y.-G. Jiang, J.~Liu, A.~Roshan~Zamir, G.~Toderici, I.~Laptev, M.~Shah, and
  R.~Sukthankar, ``{THUMOS} challenge: Action recognition with a large number
  of classes,'' 2014.

\bibitem{yeung2015every}
S.~Yeung, O.~Russakovsky, N.~Jin, M.~Andriluka, G.~Mori, and L.~Fei-Fei,
  ``Every moment counts: Dense detailed labeling of actions in complex
  videos,'' \emph{International Journal of Computer Vision (IJCV)}, 2017.

\bibitem{Krizhevsky:2017:ICD:3098997.3065386}
A.~Krizhevsky, I.~Sutskever, and G.~E. Hinton, ``Imagenet classification with
  deep convolutional neural networks,'' in \emph{NIPS}, 2012.

\bibitem{Sermanet_overfeat:integrated}
P.~Sermanet, D.~Eigen, X.~Zhang, M.~Mathieu, R.~Fergus, and Y.~Lecun,
  ``Overfeat: Integrated recognition, localization and detection using
  convolutional networks,'' \emph{ICLR}, 2014.

\bibitem{Simonyan14c}
K.~Simonyan and A.~Zisserman, ``Very deep convolutional networks for
  large-scale image recognition,'' \emph{CoRR}, 2014.

\bibitem{43022}
C.~Szegedy, W.~Liu, Y.~Jia, P.~Sermanet, S.~Reed, D.~Anguelov, D.~Erhan,
  V.~Vanhoucke, and A.~Rabinovich, ``Going deeper with convolutions,'' in
  \emph{CVPR}, 2015.

\bibitem{labelgraph}
J.~Deng, N.~Ding, Y.~Jia, A.~Frome, K.~Murphy, S.~Bengio, Y.~Li, H.~Neven, and
  H.~Adam, ``Large-scale object classification using label relation graphs,''
  in \emph{ECCV}, 2014.

\bibitem{7410495}
N.~Ding, J.~Deng, K.~P. Murphy, and H.~Neven, ``Probabilistic label relation
  graphs with ising models,'' in \emph{ICCV}, 2015.

\bibitem{inn}
H.~Hu, G.-T. Zhou, Z.~Deng, Z.~Liao, and G.~Mori, ``Learning structured
  inference neural networks with label relations,'' in \emph{CVPR}, 2016.

\bibitem{DBLP:journals/corr/NauataSM17}
N.~Nauata, J.~Smith, and G.~Mori, ``Hierarchical label inference for video
  classification,'' \emph{CoRR}, 2017.

\bibitem{lstm}
S.~Hochreiter and J.~Schmidhuber, ``Long short-term memory,'' \emph{Neural
  Computation (NC)}, vol.~9, no.~8, pp. 1735--1780, Nov. 1997.

\bibitem{NIPS2011_4212}
J.~Deng, S.~Satheesh, A.~C. Berg, and F.~Li, ``Fast and balanced: Efficient
  label tree learning for large scale object recognition,'' in \emph{NIPS},
  2011.

\bibitem{Taskar:2003:MMN:2981345.2981349}
B.~Taskar, C.~Guestrin, and D.~Koller, ``Max-margin markov networks,'' in
  \emph{NIPS}, 2003.

\bibitem{Tsochantaridis:2005:LMM:1046920.1088722}
I.~Tsochantaridis, T.~Joachims, T.~Hofmann, and Y.~Altun, ``Large margin
  methods for structured and interdependent output variables,'' \emph{Journal
  of Machine Learning Research (JMLR)}, vol.~6, pp. 1453--1484, 2005.

\bibitem{NIPS2011_4250}
K.~Grauman, F.~Sha, and S.~J. Hwang, ``Learning a tree of metrics with disjoint
  visual features,'' in \emph{NIPS}, 2011.

\bibitem{hwang2012semantic}
S.~J. Hwang, K.~Grauman, and F.~Sha, ``Semantic kernel forests from multiple
  taxonomies,'' in \emph{NIPS}, 2012.

\bibitem{Johnson2015ICCV}
J.~Johnson, L.~Ballan, and L.~Fei-Fei, ``Love thy neighbors: Image annotation
  by exploiting image metadata,'' in \emph{ICCV}, 2015.

\bibitem{conf/eccv/McAuleyL12}
J.~J. McAuley and J.~Leskovec, ``Image labeling on a network: Using
  social-network metadata for image classification.'' in \emph{ECCV}, 2012.

\bibitem{1467244}
M.~P. Kumar, P.~H.~S. Ton, and A.~Zisserman, ``Obj cut,'' in \emph{CVPR}, 2005.

\bibitem{4270067}
B.~Wu and R.~Nevatia, ``Simultaneous object detection and segmentation by
  boosting local shape feature based classifier,'' in \emph{CVPR}, 2007.

\bibitem{Kohli2008}
P.~Kohli, J.~Rihan, M.~Bray, and P.~H.~S. Torr, ``Simultaneous segmentation and
  pose estimation of humans using dynamic graph cuts,'' \emph{International
  Journal of Computer Vision (IJCV)}, vol.~79, no.~3, pp. 285--298, 2008.

\bibitem{Leibe04combinedobject}
B.~Leibe, A.~Leonardis, and B.~Schiele, ``Combined object categorization and
  segmentation with an implicit shape model,'' in \emph{ECCV Workshop}, 2004.

\bibitem{KongCVPR14}
C.~Kong, D.~Lin, M.~Bansal, R.~Urtasun, and S.~Fidler, ``What are you talking
  about? text-to-image coreference,'' in \emph{CVPR}, 2014.

\bibitem{7534740}
A.~Karpathy and L.~Fei-Fei, ``Deep visual-semantic alignments for generating
  image descriptions,'' \emph{IEEE Transactions on Pattern Analysis and Machine
  Intelligence (TPAMI)}, vol.~39, no.~4, pp. 664--676, 2017.

\bibitem{inceptionv3}
C.~Szegedy, V.~Vanhoucke, S.~Ioffe, J.~Shlens, and Z.~Wojna, ``Rethinking the
  inception architecture for computer vision,'' in \emph{CVPR}, 2016.

\bibitem{he2016deep}
K.~He, X.~Zhang, S.~Ren, and J.~Sun, ``Deep residual learning for image
  recognition,'' in \emph{CVPR}, 2016.

\bibitem{HengWang:2011:ARD:2191740.2192078}
H.~Wang, A.~Klaser, C.~Schmid, and C.-L. Liu, ``Action recognition by dense
  trajectories,'' in \emph{CVPR}, 2011.

\bibitem{6751553}
H.~Wang and C.~Schmid, ``Action recognition with improved trajectories,'' in
  \emph{ICCV}, 2013.

\bibitem{6909619}
A.~Karpathy, G.~Toderici, S.~Shetty, T.~Leung, R.~Sukthankar, and L.~Fei-Fei,
  ``Large-scale video classification with convolutional neural networks,'' in
  \emph{CVPR}, 2014.

\bibitem{6165309}
S.~Ji, W.~Xu, M.~Yang, and K.~Yu, ``3d convolutional neural networks for human
  action recognition,'' \emph{IEEE Transactions on Pattern Analysis and Machine
  Intelligence (TPAMI)}, vol.~35, no.~1, pp. 221--231, 2013.

\bibitem{10.1007/978-3-642-25446-8_4}
M.~Baccouche, F.~Mamalet, C.~Wolf, C.~Garcia, and A.~Baskurt, ``Sequential deep
  learning for human action recognition,'' in \emph{HBU}, A.~A. Salah and
  B.~Lepri, Eds., 2011.

\bibitem{beyond_horts_nippets}
J.~Y.-H. Ng, M.~Hausknecht, S.~Vijayanarasimhan, O.~Vinyals, R.~Monga, and
  G.~Toderici, ``Beyond short snippets: Deep networks for video
  classification,'' in \emph{CVPR}, 2015.

\bibitem{lrcn}
J.~Donahue, L.~A. Hendricks, M.~Rohrbach, S.~Venugopalan, S.~Guadarrama,
  K.~Saenko, and T.~Darrell, ``Long-term recurrent convolutional networks for
  visual recognition and description,'' \emph{IEEE Transactions on Pattern
  Analysis and Machine Intelligence (TPAMI)}, vol.~39, no.~4, pp. 677--691,
  2017.

\bibitem{4409011}
Y.~Ke, R.~Sukthankar, and M.~Hebert, ``Event detection in crowded videos,'' in
  \emph{ICCV}, 2007.

\bibitem{7780585}
B.~Singh, T.~K. Marks, M.~Jones, O.~Tuzel, and M.~Shao, ``A multi-stream
  bi-directional recurrent neural network for fine-grained action detection,''
  in \emph{CVPR}, 2016.

\bibitem{Montes_2016_NIPSWS}
A.~Montes, A.~Salvador, S.~Pascual, and X.~Giro-i Nieto, ``Temporal activity
  detection in untrimmed videos with recurrent neural networks,'' in \emph{NIPS
  Workshop}, 2016.

\bibitem{7780583}
S.~Ma, L.~Sigal, and S.~Sclaroff, ``Learning activity progression in lstms for
  activity detection and early detection,'' in \emph{CVPR}, 2016.

\bibitem{223161}
J.~Yamato, J.~Ohya, and K.~Ishii, ``Recognizing human action in time-sequential
  images using hidden markov model,'' in \emph{CVPR}, 1992.

\bibitem{4270156}
F.~Lv and R.~Nevatia, ``Single view human action recognition using key pose
  matching and viterbi path searching,'' in \emph{CVPR}, 2007.

\bibitem{Shi2011}
Q.~Shi, L.~Cheng, L.~Wang, and A.~Smola, ``Human action segmentation and
  recognition using discriminative semi-markov models,'' \emph{International
  Journal of Computer Vision (IJCV)}, vol.~93, no.~1, pp. 22--32, 2011.

\bibitem{NIPS2008_3449}
A.~Graves and J.~Schmidhuber, ``Offline handwriting recognition with
  multidimensional recurrent neural networks,'' in \emph{NIPS}, 2009.

\bibitem{DBLP:conf/interspeech/2014}
H.~Li, H.~M. Meng, B.~Ma, E.~Chng, and L.~Xie, Eds., \emph{Interspeech}, 2014.

\bibitem{Miller:1995:WLD:219717.219748}
G.~A. Miller, ``Wordnet: A lexical database for english,'' \emph{Communications
  of the ACM (CACM)}, vol.~38, no.~11, pp. 39--41, 1995.

\bibitem{NIPS2014_5289}
S.~J. Hwang and L.~Sigal, ``A unified semantic embedding: Relating taxonomies
  and attributes,'' in \emph{NIPS}, 2014.

\bibitem{journals/corr/GongJLTI13}
Y.~Gong, Y.~Jia, T.~Leung, A.~Toshev, and S.~Ioffe, ``Deep convolutional
  ranking for multilabel image annotation.'' \emph{CoRR}, 2013.

\bibitem{Xiao:2016:SDE:2963034.2963064}
J.~Xiao, K.~A. Ehinger, J.~Hays, A.~Torralba, and A.~Oliva, ``Sun database:
  Exploring a large collection of scene categories,'' \emph{International
  Journal of Computer Vision (IJCV)}, vol. 119, no.~1, pp. 3--22, 2016.

\bibitem{caffe}
Y.~Jia, E.~Shelhamer, J.~Donahue, S.~Karayev, J.~Long, R.~Girshick,
  S.~Guadarrama, and T.~Darrell, ``Caffe: Convolutional architecture for fast
  feature embedding,'' in \emph{ACM MM}, 2014.

\bibitem{NIPS2014_5349}
B.~Zhou, A.~Lapedriza, J.~Xiao, A.~Torralba, and A.~Oliva, ``Learning deep
  features for scene recognition using places database,'' in \emph{NIPS}, 2014.

\bibitem{pmlr-v32-donahue14}
J.~Donahue, Y.~Jia, O.~Vinyals, J.~Hoffman, N.~Zhang, E.~Tzeng, and T.~Darrell,
  ``Decaf: A deep convolutional activation feature for generic visual
  recognition,'' in \emph{ICML}, 2014.

\bibitem{audio}
S.~Hershey, S.~Chaudhuri, D.~P.~W. Ellis, J.~F. Gemmeke, A.~Jansen, C.~Moore,
  M.~Plakal, D.~Platt, R.~A. Saurous, B.~Seybold, M.~Slaney, R.~Weiss, and
  K.~Wilson, ``Cnn architectures for large-scale audio classification,'' in
  \emph{ICASSP}, 2017.

\bibitem{adam}
D.~P. Kingma and J.~Ba, ``Adam: {A} method for stochastic optimization,''
  \emph{CoRR}, 2014.

\bibitem{ucf101}
K.~Soomro, A.~Roshan~Zamir, and M.~Shah, ``{UCF101}: A dataset of 101 human
  actions classes from videos in the wild,'' in \emph{CRCV-TR-12-01}, 2012.

\bibitem{Patwardhan2003IncorporatingDA}
S.~V. Patwardhan, ``Incorporating dictionary and corpus information into a
  context vector measure of semantic relatedness,'' Master's thesis, University
  of Minnesota, 2003.

\end{thebibliography}
\end{document}